\documentclass[]{llncs}
\usepackage{graphicx}

\usepackage{tikz}
\usepackage{comment}
\usepackage{amsmath,amssymb} 
\usepackage{color}

\usepackage[accsupp]{axessibility}  
\usepackage[
 font=tiny 
 ,farskip=0pt
 ]{subfig}

\usepackage{graphicx}
\usepackage{amsmath}
\usepackage{amssymb}
\usepackage{booktabs}
\usepackage{multirow}
\usepackage{adjustbox}
\usepackage{threeparttable}
\usepackage{multicol}
\usepackage[font=footnotesize,skip=0pt]{caption}
\usepackage{xcolor}
\usepackage{enumitem}
\usepackage{breakcites}
\usepackage[pagebackref,breaklinks,colorlinks]{hyperref}
\usepackage{floatrow}

\definecolor{mycolor}{HTML}{F7F8E0}
\definecolor{darkcyan}{RGB}{0,113,194}
\definecolor{forestgreen}{RGB}{34,139,34}
\definecolor{customyellow}{RGB}{255,217,102}
\definecolor{mypink3}{cmyk}{0, 0.7808, 0.4429, 0.1412}

\newcommand*{\modelname}{YORO }
\newcommand*{\tasknameshort}{VG }
\newcommand*{\tasknamelong}{Visual Grounding }

\begin{document}
\pagestyle{headings}
\mainmatter

\title{YORO - Lightweight End to End Visual Grounding} 



\titlerunning{YORO - Lightweight End to End Visual Grounding}
%
\author{Chih-Hui Ho\inst{2}\thanks{Work done during internship at Amazon.} \and
Srikar Appalaraju\inst{1}\thanks{Corresponding author.} \and
Bhavan Jasani \inst{1} \and
R. Manmatha \inst{1} \and
Nuno Vasconcelos \inst{2}}
%

%
\institute{AWS AI Labs  \email{\{srikara,bjasani,manmatha\}@amazon.com }
 \and
UC San Diego  \email{\{chh279,nuno\}@ucsd.edu}
}

\maketitle

\begin{abstract}
We present YORO - a multi-modal transformer encoder-only architecture for the Visual Grounding (VG) task. This task involves localizing, in an image, an object referred via natural language. Unlike the recent trend in the literature of using multi-stage approaches that sacrifice speed for accuracy, YORO seeks a better trade-off between speed an accuracy by embracing a single-stage design, without CNN backbone. YORO consumes natural language queries, image patches, and learnable detection tokens and predicts coordinates of the referred object, using a single transformer encoder. To assist the alignment between text and visual objects, a novel patch-text alignment loss is proposed. Extensive experiments are conducted on 5 different datasets with ablations on architecture design choices. YORO is shown to support real-time inference and outperform all approaches in this class (single-stage methods)  by large margins. It is also the fastest VG model and achieves the best speed/accuracy trade-off in the literature.
\end{abstract}

\vspace{-10pt}
\section{Introduction}
\begin{figure}[t]\RawFloats
\centering
\begin{tabular}{cc}
  \includegraphics[width=0.46\linewidth]{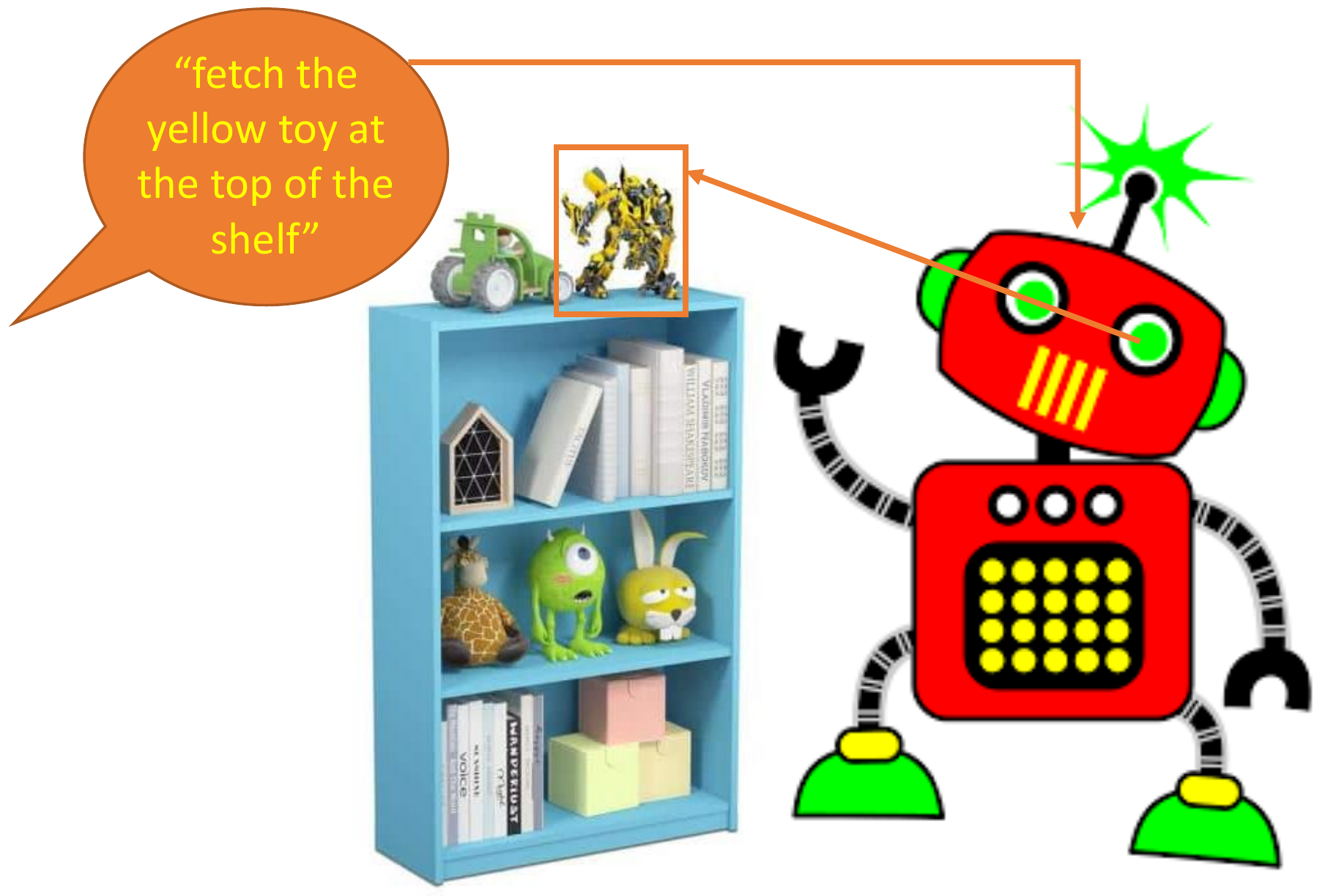}  &  \includegraphics[width=0.45\linewidth]{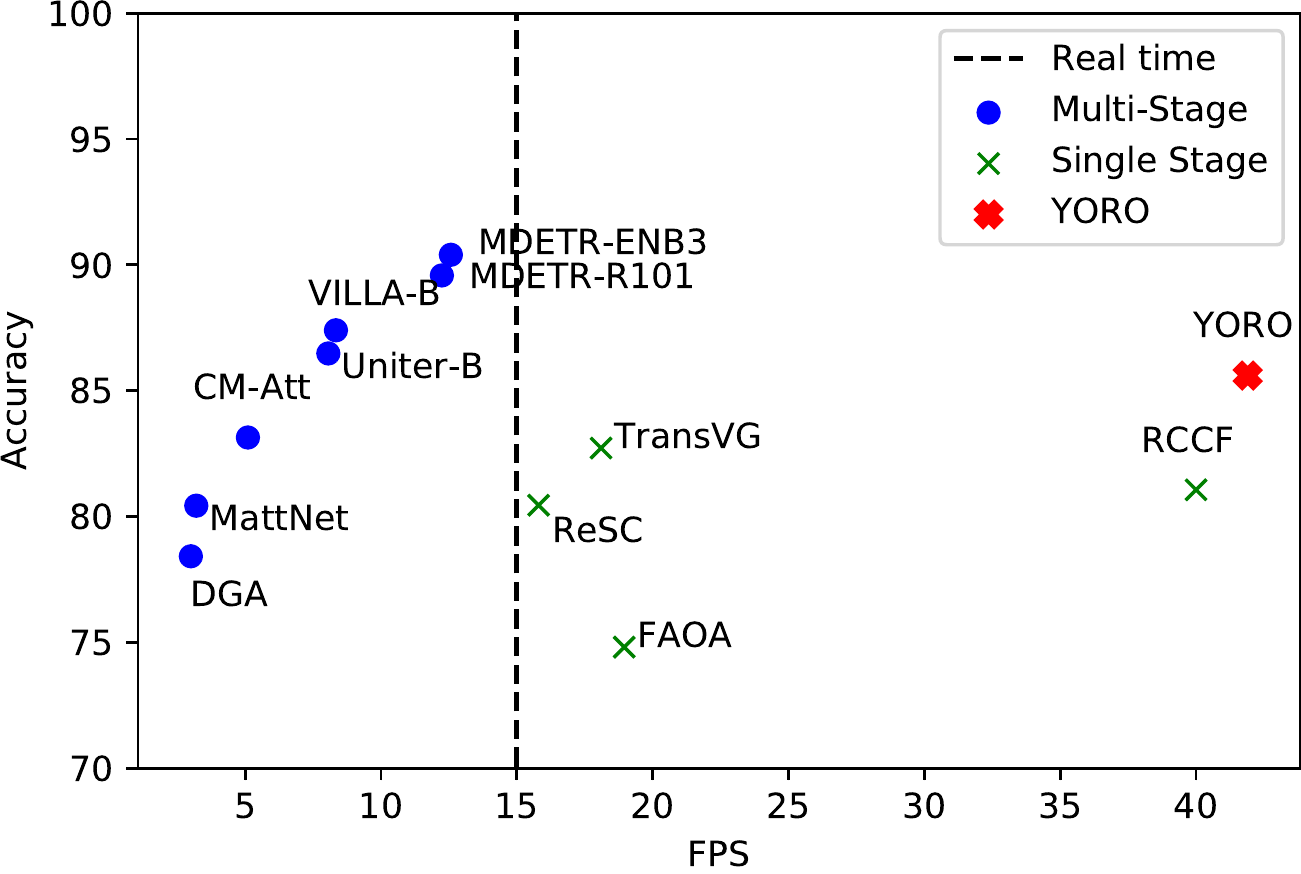}\\
  \scriptsize{(a)} & \scriptsize(b) 
\end{tabular}
\caption{
(a) Example of VG with limited computational resources.
(b) \textbf{Accuracy-Speed plot}: \modelname surpasses real time FPS 15 and achieves better trade-off in the literature.}
\label{fig:speed_acc_plot_splash}
\end{figure}

\vspace{-5pt}
There has been significant recent interest  in \tasknamelong (VG) ~\cite{mattnet,Yu2017AJS,Luo_2017_CVPR,Deng_2018_CVPR,Zhang_2018_CVPR,Wang_2019_CVPR,Yang_2019_ICCV,Liu_2019_CVPR,Zhuang_2018_CVPR,Liu_2019_ICCV,Yu2018RethinkingDA,Yang_2019_ICCV_DGA,Yang_2019_CVPR}. This aims to localize, in an image, an object referred to by natural language, using a text query (see Fig. \ref{fig:speed_acc_plot_splash}(a)). VG is potentially useful for many applications, ranging from cloud-based image retrieval from large corpora to resource constrained problems on devices of limited computational capacity. For example, in Fig.~\ref{fig:speed_acc_plot_splash} (a), a robot must use its limited computing resources to resolve a natural language query and locate the desired object. These applications currently pose an extreme challenge to VG, by compounding the difficulty of the task itself with the need to solve it efficiently. When this is the case, computational efficiency becomes an unavoidable requirement of architecture design. Several examples exist in the vision literature, where branches of computationally efficient architectures have evolved for problems like object recognition~\cite{mobilenetv1} or detection~\cite{yolo}, among others.  The goal is not necessarily to develop the most powerful method in the literature, but to find the best trade-off~\cite{Huang2017SpeedAccuracyTF} between task performance and some variable that reflects computation, e.g. speed or latency. This has led to the introduction of architectures, such as the MobileNet family~\cite{mobilenetv1,mobilenetv2,mobilenetv3} for recognition or the YOLO family~\cite{yolo,yolov2,yolov3,yolov4} for object detection, that have become popular despite their weaker than state of the art accuracy. Given the obvious relationship between object detection and VG, it is not surprising that similar trends are starting to emerge for the latter. Like object detectors, VG methods can be broadly grouped in two categories:  single ~\cite{Yang_2019_ICCV,rccf,ssg,mcn,resc,transvg} and multi-stage~\cite{uniter,mattnet,Liu_2019_CVPR,vlbert,villa} methods. Multi-stage methods typically rely on a pre-trained visual backbone (such as the FasterRCNN~\cite{fasterrcnn}) as a visual encoder module. This simplifies the VG problem, reducing it to the selection of the object proposals that best match the query text. Like multi-stage object detectors, they are more accurate than single stage methods, but usually too complex for the applications of Fig.~\ref{fig:speed_acc_plot_splash} (a). In fact, the run-time complexity of many of these VG systems is lower-bounded by the run-time computation of the FasterRCNN~\cite{fasterrcnn} proposal stage, which is usually considered too expensive even for object detection, in domains like robotics. In these domains, single-stage detectors such as YOLO~\cite{yolo,yolov2,yolov3,yolov4} are the architecture of choice. Unsurprisingly, the YOLO family has been the backbone of choice for various single-stage VG systems~\cite{ssg,Yang_2019_ICCV}. However, these and the other CNN-based  single stage VG systems~\cite{Yang_2019_ICCV,rccf,ssg,mcn,resc} previously proposed in the literature lack the powerful attention mechanisms that enable recent transformer architectures to excel at multi-modal data fusion~\cite{transvg}. This usually results in weak accuracy.


In this work, we seek to endow single stage VG methods with transformer-style attention, so as to enable a better trade-off between accuracy and speed than currently available methods. 
For this, we propose an efficient and end-to-end differentiable VG architecture, denoted as{\it You Only Refer Once\/} (YORO). YORO is inspired by a transformer architecture, ViT~\cite{vit}, that has achieved success for image classification. ViT has also recently been extended to a transformer architecture,  ViLT~\cite{vilt}, for vision-language tasks such as visual reasoning~\cite{NLVR2} or visual question answering~\cite{vqav2}. 
We investigate the more challenging extension of ViLT for tasks that require object localization, such as VG. The main challenge is how to establish explicit connections between words and bounding boxes without the addition of an object detector. Inspired by recent approaches to the design of transformers for object detection~\cite{mdetr,detr,yolos}, we augment the ViLT architecture with a set of detection tokens that enable this localization. This makes \modelname an encoder-only multi-modal transformer architecture that consumes three data streams: referred text tokens, an image in the form of flattened patches, and learnable detection tokens. \modelname then uses self-attention to fuse information from all these streams in order to localize the referred object. 

Similarly to transformer-based detectors~\cite{detr,yolos}, \modelname is trained by Hungarian matching between the localization hypotheses derived from its tokens and the ground-truth bounding boxes. This combines a bounding box regression loss that minimizes the difference between predicted and ground truth boxes and a classification loss that associates the referred text to the bounding box. The alignment between language and vision representations is further strengthened by an object-text alignment loss that operates in feature space, encouraging the visual features of the predicted object to match the text features of the corresponding text. However, this must be done carefully. In the early stages of training,  when object predictions tend to be incorrect, this object-text alignment loss can hamper learning. To both address this issue and encourage more fine-grain alignment, we propose a novel patch-text alignment loss that aligns features of input patches to their corresponding text. Extensive experiments on five datasets show that \modelname establishes the new SOTA for single stage VG models.  
Overall, as illustrated in Fig.~\ref{fig:speed_acc_plot_splash}(b), \modelname (red) achieves the best accuracy/speed trade-off among the methods in the literature. It has substantially better accuracy than previous single stage VG methods (green) and is substantially faster than  multi-stage VG models (blue). Ablation studies compare different variants of the proposed model and demonstrate the effectiveness of the proposed patch-text alignment loss.  With this we summarize our contributions:
\begin{itemize}[leftmargin=*]
\setlength\itemsep{1pt}
	\item A novel multi-modal visual grounding architecture \modelname is proposed using an encoder-only transformer (Sec. \ref{sec:architecture}). \modelname is a simple single-stage design and is end-to-end trainable. 
	\item A novel patch-text alignment loss is introduced for enabling fine-grained visual language alignment (Sec. \ref{sec:PTLoss}). 
	\item \modelname achieves the best accuracy/speed trade-off in the literature, outperforming existing real-time VG methods on five datasets (Sec. \ref{sec:experiments}).
\end{itemize}


\vspace{-5pt}
\section{Related work}\vspace{-5pt}
Prior work in visual grounding (VG) may be mainly classified into multi-stage and single-stage approaches based on the design of the visual branch. 

\vspace{.05in}
\noindent\textbf{Multi-stage VG approaches}~\cite{uniter,villa,vilbert,12in1,vgtr,vlt5,slr,mattnet} usually leverage a region proposal network (RPN) ~\cite{fasterrcnn} to handle the selection of candidate object locations. In a first-stage, the RPN proposes a set of object candidates. In a second-stage, the \tasknameshort model selects the region proposal that best matches the query text. This reduces \tasknameshort to a retrieval problem~\cite{cmn,hu2016natural,groundr,mcb}, where the model only needs to search within the proposed object candidates. Similarly to two-stage object detectors~\cite{Liu2019DeepLF, Jiao2019}, these methods are powerful for VG. However, like two-stage detectors, they tend to be computationally heavy, making them impractical for real-time operation on complexity constrained devices, as illustrated in Fig.~\ref{fig:speed_acc_plot_splash}.  When compared to the object detection literature~\cite{Liu2019DeepLF, Jiao2019}, the optimization of the trade-off between speed and performance has received  much less attention in VG. On the contrary, recent VG methods~\cite{vlt5,mdetr} pursue higher accuracy by adopting more powerful language models~\cite{t5,roberta} and encoder-decoders transformer architectures~\cite{t5,transformer,prophetnet,bigbird,zhang2019pegasus,Raffel2020ExploringTL,bart} with more parameters and slower speeds.  Instead, \modelname pursues a better trade-off between accuracy and speed. In this context, it is not clear that a multi-stage architecture is the best solution. Since there is no way to recover objects missed by the RPN, strong performance can require a large number of proposals~\cite{Yu2018RethinkingDA,Nagaraja2016ModelingCB}, which is inherently expensive. In fact, the use of a preliminary object selection prevents the exploitation of contextual clues that may be available in the query text (e.g. the component ``on the top of the shelf'' of the query of Fig.~\ref{fig:speed_acc_plot_splash}) and could be critical for solving the task with less proposals. The integration of language and vision is, after all, the difference between object detection and VG. This suggests that single stage architectures, like YORO, could be more effective when computation has a cost. Our experimental results support this hypothesis. 

\vspace{.05in}
\noindent\textbf{Single-stage VG approaches}~\cite{Yang_2019_ICCV,rccf,ssg,mcn,resc,transvg}
achieve greater inference speeds by discarding the proposal generation stage. Most of these methods are based on CNN object detectors, namely the single-stage detector YOLOV3~\cite{yolov3}. This is, for example, adopted to extract visual features by SSG~\cite{ssg} and FAOA~\cite{Yang_2019_ICCV}. These features are then fused with text features to regress the bounding boxes. ReSC~\cite{resc} improves FAOA~\cite{Yang_2019_ICCV} with a recursive sub-query construction module. MCN~\cite{mcn} further extends YOLOV3~\cite{yolov3} for the joint solution of VG and segmentation, using multitask losses. These models lack the powerful attention mechanisms of the transformer architecture, which are critical for the integration of information both spatially, across image regions, and modally, across the vision and language information streams. When compared to them, \modelname has the benefit of leveraging the transformer to implement this powerful integration of information. This enables substantially higher accuracy than previous single-stage approaches.  Recently, TransVG~\cite{transvg} has also adopted a transformer model~\cite{transformer} to fuse language and vision features. 
While TransVG does not use a two-stage object detector, it uses two stages of transformers. The vision and language streams are first processed by independent transformers, whose outputs are then fed to a multi-modal transformer that integrates the two modalities. This makes it substantially slower than YORO, which performs all operations with a single transformer, and could give rise to loss of information needed to reason jointly about vision and language, as discussed above for the RPN. The fact that TransVG is both slower and less accurate than YOLO suggests that there is no benefit to the two transformer stage architecture. 

\vspace{.1in}
\noindent{\bf Vision Transformer.}
YORO is based on the Vision transformer (ViT), introduced in~\cite{vit} for image classification by direct consumption of image patches, i.e. without a preliminary CNN backbone. Beyond classification~\cite{vit,clip,deit}, the vision transformer has been applied to detection~\cite{detr,zhu2021deformable,yolos} and segmentation~\cite{Ranftl2021,Strudel_2021_ICCV}, 
as discussed in the recent surveys of~\cite{Khan2021TransformersIV,Han2020ASO}. 
Recently, ViLT~\cite{vilt} has extended the ViT for various visual-language tasks that do not require object localization, such as visual question answering.  
The extension to tasks, such as VG, that require object localization is non-trivial. This is due to the need to establish explicit connections between words and bounding boxes, which requires much more fine-grained understanding of both language and images. YORO addresses this limitation by the addition of learnable detection tokens and Hungarian matching between token-based object predictions and ground truth bounding boxes, a component of modern transformer-based object detectors~\cite{detr,zhu2021deformable,yolos}. 
To the best of our knowledge, it is the first architecture to perform \tasknameshort tasks with an encoder-only transformer without a visual backbone. This is critical for speed. YORO is shown to achieve the best trade-off between speed and accuracy in the VG literature, as illustrated in Fig.~\ref{fig:speed_acc_plot_splash} (b).


\vspace{-5pt}
\section{\modelname Architecture}\vspace{-5pt}
\label{sec:architecture}
The \modelname architecture is summarized in Fig.~\ref{fig:arch}. The input query sentence and image are first converted into text and visual embeddings respectively. These embeddings are augmented by a set of learnable detection tokens and fed into a transformer encoder. Within the transformer, learning happens via self-attention \cite{transformer}. The transformer predicted features corresponding to the detection tokens are finally fed into several heads, implemented with a multi-layer perceptron, which produce class and bounding box predictions. The model is trained end-to-end, using a combination of losses.
Each model component is discussed next. 

\vspace{-5pt}
\subsection{Multi-Modal Inputs}\label{sec:trans_input}\vspace{-5pt}

YORO consumes multi-modal inputs: language, vision and detection tokens.

\begin{figure}[t]\RawFloats
\centering
\begin{tabular}{c}
    \includegraphics[width=0.98\linewidth]{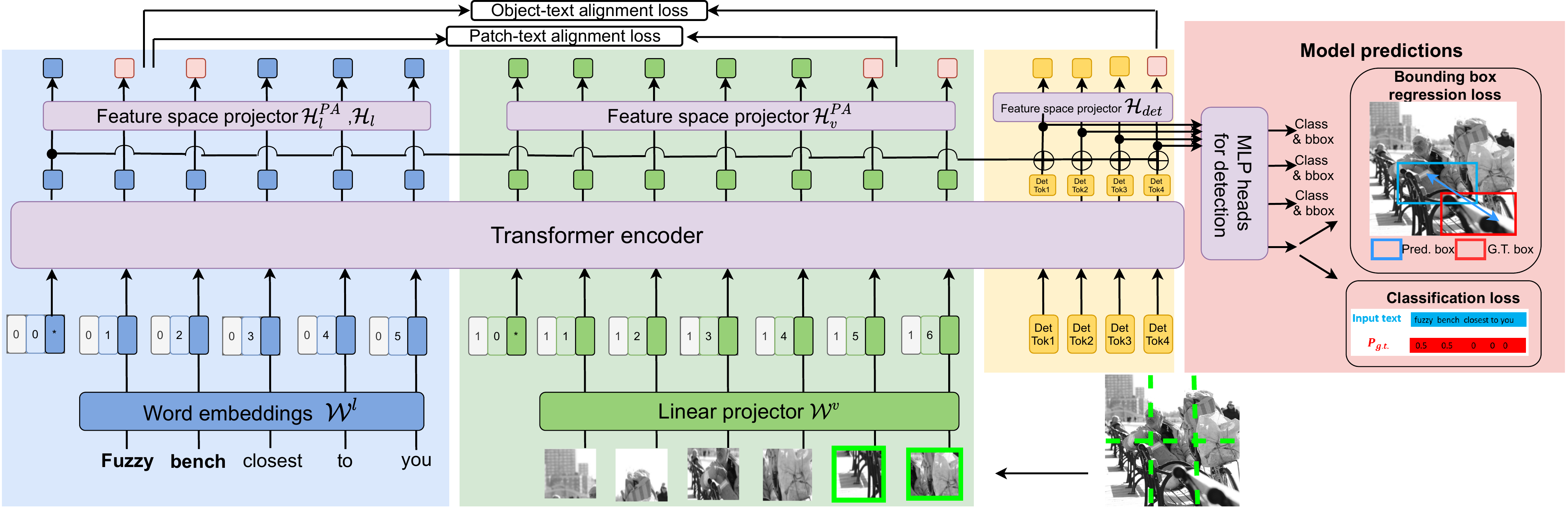} 
\end{tabular}
\caption{\textbf{\modelname architecture}. \textcolor{blue}{Blue}, \textcolor{forestgreen}{green}, \textcolor{customyellow}{yellow} and \textcolor{mypink3}{pink} blocks represent language, vision, detection and prediction branches respectively. YORO does \textbf{not} use a large pre-trained visual backbone. Input image is divided into patches. Those of IOU$\geq0.5$ with ground truth box (red) are marked in light green.}
\label{fig:arch}
\end{figure}

\noindent\textbf{The Language Modality} is depicted in the blue block of Fig.~\ref{fig:arch}. The referred text is converted into sequences of tokens with a BERT tokenizer~\cite{bert}. Assume the sequence has length $m$ and a dictionary size of $v$ for the pre-trained tokenizer. Each text token $t_i \in \mathbb{R}^{v}$ in the sequence $\{t_1 ; \dots ; t_m\}$ is then projected into a $d$ dimensional feature vector $f^l_i=\mathcal{W}^lt_i$, where $\mathcal{W}^l \in \mathbb{R}^{d\times v}$ is a projection layer. The $m$ projected token vectors are then augmented with a text classification token $f_{cls}^l \in \mathbb{R}^{d}$. Similar to \cite{vilt}, each of these tokens is summed with a text positional embedding $f_{pos}^l \in \mathbb{R}^{(m+1)\times d}$ and a modality type embedding $f_{type}^l\in \mathbb{R}^{(m+1)\times d}$. The positional embedding specifies the location of each token, while the modality type embedding tells apart the text input from visual input. The overall text input embedding may be written as
\begin{equation}
    f^l = [f_{cls}^l ; f^l_1 ; \dots ;f^l_m] + f_{pos}^l +  f_{type}^l.
\end{equation}

\noindent\textbf{The Vision Modality} is shown in the green block of Fig.~\ref{fig:arch}. The input image $I\in \mathbb{R}^{H\times W \times 3}$ is partitioned into $n$ visual patches $\{p_1, \dots, p_n\}$ of size $s$, where $p_i \in \mathbb{R}^{s^2 \times 3}$ and $n=\frac{HW}{s^2}$. Each visual patch is then projected, with a linear layer of weight matrix $\mathcal{W}^v \in \mathbb{R}^{d\times (s^2 \times 3)}$, into a $d$ dimensional visual patch feature $f^v_i=\mathcal{W}^vp_i$. Similar to the language input, the $n$ visual patch features are augmented with a visual classification token $f_{cls}^v \in \mathbb{R}^{d}$ and summed with a visual positional embedding $f_{pos}^v \in \mathbb{R}^{(n+1)\times d}$ and a modality type embedding $f_{type}^v\in \mathbb{R}^{(n+1)\times d}$, to produce an overall  visual input embedding
\begin{equation}
    f^v = [f_{cls}^v ; f^v_1 ; \dots ;f^v_n] + f_{pos}^v +  f_{type}^v.
\end{equation}

\noindent\textbf{Learnable Detection Tokens}. The vision and language  models above are similar to those implemented by ViT~\cite{vit} and ViLT~\cite{vilt}. These models were proposed for tasks, such as image classification or visual question-answering that do not require object localization. In this case, the use of holistic classifications tokens $f_{cls}^l,f_{cls}^v$ is sufficient to encode the image identity. However, our experiments (see ablations in Sec.~\ref{sec:ablation}) show that this is not the case for VG, which depends critically on object localization. In the object detection literature, this problem is addressed by introducing learnable detection tokens, which are individually associated with bounding boxes~\cite{detr}. \modelname leverages such tokens, as shown in the yellow block of Fig.~\ref{fig:arch}. The text $f^l$ and visual $f^v$ embeddings are complemented by $q$ learnable detection tokens $f^{det} = [f^{det}_1 ; \ldots ; f^{det}_q]$. Each token represents an object and different tokens may represent objects of different sizes.

\vspace{-5pt}
\subsection{Transformer Encoder}\label{sec:trans_encoder}\vspace{-5pt}
\modelname transformer encoder consists of $D$ stacked layers, each including a multi-headed self-attention (MSA) network and a feed forward network (FFN). Denoting $x_d$ as the input of layer $d$, the transformer output is computed as follows:
\begin{eqnarray}
   x_0 &=& [ f^l; f^v ; f^{det} ] \\
    x'_d &=& x_{d-1} + \text{MSA}(\text{LN}(x_{d-1})),  \quad  d=1\dots D \\
  x_{d} &=& x'_d + \text{FFN}(\text{LN}(x'_d))), \quad  d=1\dots D
\end{eqnarray}
where LN denotes layer normalization~\cite{transformer}. $D$ is set to 12 in our experiments.

\vspace{-5pt}
\subsection{Multi-Modal Transformer Outputs}\label{sec:trans_output}\vspace{-5pt}
The transformer input $[f^l ; f^v ; f^{det}]$ is a sequence of length $n+m+q+2$. The output is a sequence $[o^l ; o^v ; o^{det}]$ of the same length, where $o^l= [o^l_{cls} ; o^l_{1} ; \dots ; o^l_{m}]$, $o^v= [o^v_{cls} ; o^v_{1} ; \dots ; o^v_{n}]$ and $o^{det}= [o^{det}_{1} ; \dots ; o^{det}_{q}]$, are the outputs of the language, vision and detection branches respectively. The transformed text class token $o^l_{cls}$ is added to the transformed detection tokens $o^{det}_{i}$ before being passed to the detection heads. This strengthens the dependence of the bounding box predictions on the input text, beyond what would be possible by the use of attention alone.

\vspace{-5pt}
\subsection{Feature Projector and Detection Heads}\label{sec:det_heads} \vspace{-5pt}
To optimize \modelname with the proposed losses, several modules are added to the transformer output, including a detection head for the bounding box and classification loss (red block in Fig. ~\ref{fig:arch}), and a projection head to enable the object-text alignment and the patch-text alignment loss discussed in Sec.~\ref{sec:PTLoss}. All these heads are stacked fully connected layers. 
We next discuss the details of all losses.

\vspace{-5pt}
\section{Training Losses}\label{sec:losses}\vspace{-5pt}
\modelname is trained with 4 different losses: (1) a bounding box regression loss to regress bounding box coordinates, (2) a classification loss to classify which text tokens the bounding box is associated to, (3) an object-text alignment loss that aligns detection token features with corresponding text features and (4) a novel  patch-text alignment loss that aligns image patches and text tokens. During training, Hungarian matching~\cite{Kuhn1955Hungarian} computes the optimal bipartite matching between predicted and ground truth boxes. During inference, the bounding box of highest classification score is selected. 

\vspace{.05in}
\noindent{\bf Bounding Box Regression Loss.}
Two losses are used for bounding box regression: the $L_1$ and the generalized IoU loss~\cite{giou} $L_{giou}$. These losses are applied to ground truth bounding box $B_{g.t.}\in \mathbb{R}^4$ and predicted box $B_{pred}\in \mathbb{R}^4$ pairs. This results in bounding box regression loss
\begin{equation}
    L^{bbox}=\lambda_1 L_1(B_{pred}, B_{g.t.}) + \lambda_2 L_{giou}(B_{pred}, B_{g.t.}),
\end{equation}
where $\lambda_1$ and $\lambda_2$ are experimentally set to 2 and 5 respectively.

\vspace{.05in}
\noindent{\bf Classification Loss.}
Unlike the classification losses of object detection, where a single object class is predicted, \modelname predicts the set of language tokens to be associated with a bounding box. This is needed to encourage the association of the text phrase with the corresponding objects. For example, in Fig.~\ref{fig:arch} the red box in the image is associated with the words ``fuzzy'' and ``bench'' with probabilities 0.5 and 0.5, respectively. This is implemented by generating predictions with a softmax output of dimension  equal to the maximum token length and using a soft cross entropy loss $L_{sce}$ based on the ground-truth text/object assignments. This is the classification loss $L^{cls}=L_{sce}(P_{pred}, P_{g.t.})$, where $P_{pred}$ is the predicted probability distribution and $P_{g.t.}$ the ground truth distribution.

\vspace{.05in}
\noindent{\bf Object-Text Alignment Loss.}\label{sec:OTLoss}
Inspired by recent uses of region-text alignment for fusing words and image regions~\cite{mdetr,uniter,li2019vsrn,Diao2021SGRAF,lee2018stacked,chen2020adaptive}, \modelname further aligns the language transformer output $o^l$ and the detection transformer output $o^{det}$ in a suitable feature space. 
$o^l$ and $o^{det}$ are first mapped to a common feature space $\cal F$, with projection heads $\mathcal{H}_l$ and $\mathcal{H}_{det}$, which implement the mappings $\hat{o}^l = \mathcal{H}_l(o^l)$ and $\hat{o}^{det} = \mathcal{H}_{det}(o^{det})$, where  $\mathcal{H}_{det}$, $\mathcal{H}_{det}$ are linear projections.  Consider an image with $g$ ground truth bounding boxes and let $\mathcal{T}_i \subseteq \{j : 1 \leq j \leq m\}$, where $m$ is the length of the language token sequence,  be the subset of language token indices that a projected detection feature $\hat{o}^{det}_i$ should be aligned to. This alignment is encouraged with the object-to-text alignment (OTA) loss 
\begin{equation}
L^{OTA}=\sum_{i=1}^{g}\frac{1}{|\mathcal{T}_i|}\sum_{j\in \mathcal{T}_i} -\log\left(\frac{\exp(\hat{o}^{{det}^\intercal}_i\hat{o}^l_j/\tau)}{\sum_{k=1}^{m}\exp(\hat{o}^{{det}^\intercal}_i\hat{o}^l_k/\tau)}\right),
\end{equation}
where $\tau$ is experimentally set to 0.07 . Consider Fig.~\ref{fig:arch} and assume, for example, that detection token $4$ is selected to match the groundtruth (red) bounding box by the Hungarian matching algorithm. In this case, $L^{OTA}$ encourages the embeddings $\hat{o}^l_0$ and $\hat{o}^l_1$ of the words ``fuzzy'' and ``bench'' to be closer to the embedding of $\hat{o}^{det}_4$ than those of the words ``closest,'' ``to,'' and ``you''. 

Similarly, let $\mathcal{O}_i \subseteq \{j : 1 \leq j \leq g\}$ be the set of object indices that a projected language token feature $\hat{o}^{l}_i$ should be aligned to. This alignment is encouraged by the text-object alignment loss $L^{TOA}$
\begin{equation}
L^{TOA}=\sum_{i=1}^{m}\frac{1}{|\mathcal{O}_i|}\sum_{j\in \mathcal{O}_i} -\log\left(\frac{\exp(\hat{o}^{l^\intercal}_i\hat{o}^{det}_j/\tau)}{\sum_{k=1}^{g}\exp(\hat{o}^{l^\intercal}_i\hat{o}^{det}_k/\tau)}\right),
\end{equation}
which is a symmetric version of $L^{OTA}$. The overall object alignment loss is $L^{OA} = \frac{1}{2}L^{OTA} + \frac{1}{2}L^{TOA}$.

\begin{figure}\RawFloats
\centering
\begin{tabular}{ccc}
    \includegraphics[width=0.27\linewidth]{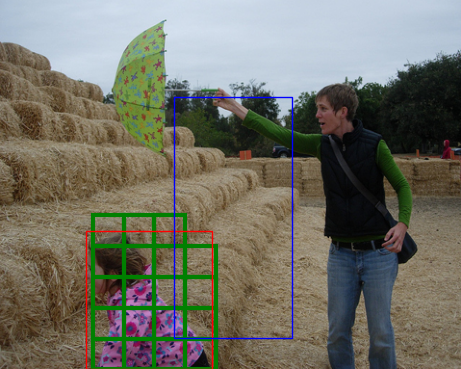} & 
    \includegraphics[width=0.27\linewidth]{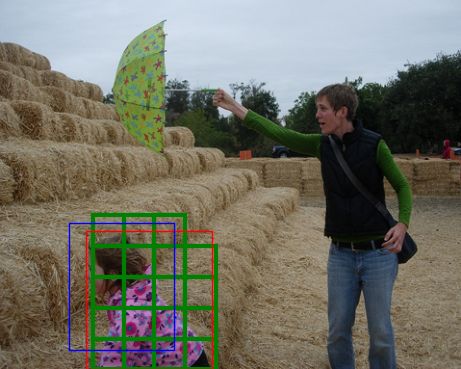} &
    \includegraphics[width=0.27\linewidth]{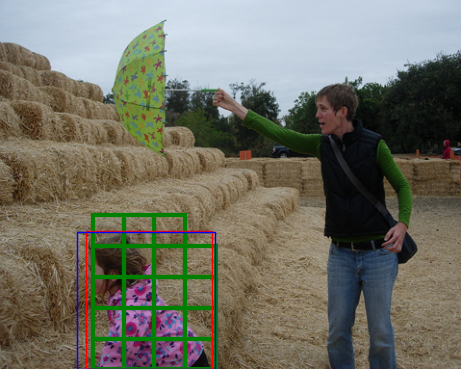} 
    \\
    \scriptsize{Epoch 0} & \scriptsize{Epoch 5} & \scriptsize{Epoch 20}
    \\
\end{tabular}
\caption{\textbf{Visualization of predicted box} through training epochs. \textcolor{red}{Red}, \textcolor{forestgreen}{green} and \textcolor{blue}{blue} boxes indicate ground truth box, ground truth patch and predicted box respectively. Patch-Text Alignment loss helps localization. Best viewed in color digitally.}
\label{fig:pa_loss_vis}
\end{figure}

\vspace{.05in}
\noindent{\bf Patch-Text Alignment Loss.}\label{sec:PTLoss}
While the object-text alignment loss is desirable to encourage feature space alignment between {\it predicted\/} boxes and corresponding text tokens, this assumes that the predicted boxes are correct.
However, as illustrated in Fig.~\ref{fig:pa_loss_vis}, this assumption is frequently violated during training, especially in the early epochs. In this case, 
the object-text alignment loss can provide ambiguous supervision (e.g. supervision from background instead of the object in the early epochs of Fig.~\ref{fig:pa_loss_vis}). To address this limitation, we propose to align the features extracted from the {\it ground truth patches at the input of the network\/} (green patches of Fig.~\ref{fig:arch}) with the features of the associated text tokens. More specifically, the indices of input patches ($p_5$ and $p_6$ in Fig.~\ref{fig:arch}) that have at least $0.5$ IOU with the ground truth box (red box in Fig.~\ref{fig:arch}) are first identified and the corresponding features aligned with those of the ground truth language tokens (``fuzzy'' and ``bench'' in Fig.~\ref{fig:arch}). As shown in Fig.~\ref{fig:pa_loss_vis}, while each patch feature may not provide full information about the object, the patch features never provide contradictory supervision. This helps stabilize the learning.  

\begin{figure}\RawFloats
\begin{tabular}{ccc}
\centering
    \includegraphics[width=0.94\linewidth]{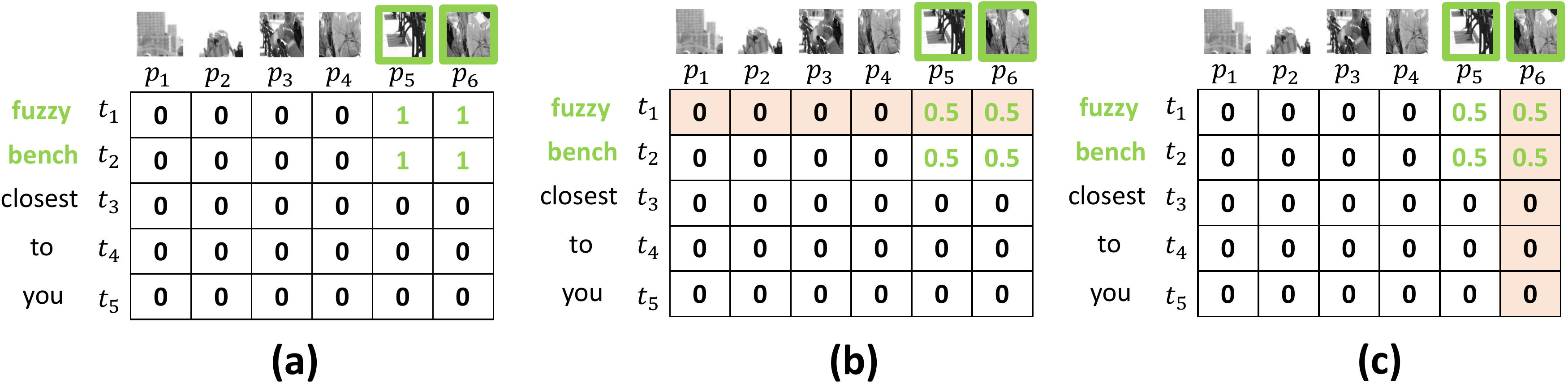} 
\end{tabular}
\caption{\textbf{Patch-Text Alignment Loss}: (a) Ground truth for patch-text alignment. (b) Column wise normalization (c) Row wise normalization.}
\label{fig:pa_groundtruth}
\end{figure}

To align text and patch features, the text transformer output $o^l$ and vision transformer outputs $o^v$ are first projected to a shared feature space using the patch alignment projection heads $\mathcal{H}^{PA}_l$ and $\mathcal{H}^{PA}_{v}$ respectively, which leads to $\Tilde{o}^l$ and $\Tilde{o}^{v}$. Given $m$ text tokens and $n$ visual patches, a ground truth table $A \in \mathbb{R}^{m \times n}$ is then created, where $A_{ij}$ is 1 if text token $i$ and visual patch $j$ should be aligned and 0 otherwise, as shown in Fig.~\ref{fig:pa_groundtruth}(a). This matrix is column normalized into matrix $A^C$ (Fig.~\ref{fig:pa_groundtruth} (b)) using 
\begin{equation}
A^c_{ij} = \left\{ 
  \begin{array}{ c l }
    \frac{A_{ij}}{\sum_i A_{ij}} & \quad \textrm{if } i \in \mathcal{S}^{TPA} \\
    0                 & \quad \textrm{otherwise}
  \end{array},
\right.    
\label{eq:pa_loss_ac}
\end{equation}
where $\mathcal{S}^{TPA}=\{i|\sum_i A_{ij} > 0\}$ is the set of token indices associated with at least 1 patch. For example, $\mathcal{S}^{TPA} = \{1,2\}$ for Fig.~\ref{fig:pa_groundtruth}. The row-wise normalization of $A$ into matrix $A^r$ is defined in a similar fashion, as shown in Fig.~\ref{fig:pa_groundtruth} (c).

To encourage the fine-grained alignment between patches and text, a text-patch alignment loss $L^{TPA}$ is defined as
\begin{equation}
    L^{TPA} = \sum_i  \sum_j p^c_{ij} ln \frac{p^c_{ij}}{A^c_{ij}} \quad \forall i \in \mathcal{S}^{TPA} ,
    \label{eq:patch_alignment_text_to_patch}
\end{equation} 
where $p^{c}_{ij}=\frac{exp(\Tilde{o}^{l^\intercal}_i\Tilde{o}^v_j)/\tau}{\sum_k exp(\Tilde{o}^{l^\intercal}_i\Tilde{o}^v_k)/\tau}$ is a contrastive distribution defined over the language and visual embedding.
For example,  the word ``fuzzy'' ($t_1$) of Fig.~\ref{fig:pa_groundtruth}(b) should be aligned with patch $p_5$ and $p_6$ according to the ground truth distribution $A^c_1=[0; 0; 0; 0; 0.5; 0.5]$. (\ref{eq:patch_alignment_text_to_patch}) minimizes the KL divergence between this and the predicted distribution $p^{c}_{1}=[p^{c}_{11};\dots;p^{c}_{1n}]$. This encourages the alignment of the feature vector $\Tilde{o}_1^{l}$ extracted from word $t_1$ with the feature vectors $\Tilde{o}_5^v$ and $\Tilde{o}_6^v$ extracted from patches $p_5$ and $p_6$. 

Similar to the $L^{TPA}$ of (\ref{eq:patch_alignment_text_to_patch}), the patch-text alignment loss $L^{PTA}$ aligns each patch to its associated tokens. As shown in Fig.~\ref{fig:pa_groundtruth}(c), $p_6$ should be aligned to $t_1$ and $t_2$. The final patch alignment loss $L^{PA}$ is $L^{PA}=\frac{1}{2}L^{TPA} + \frac{1}{2}L^{PTA}$.

\vspace{.05in}
\noindent{\bf Combined Loss}
The final loss combines the bounding box regression loss, the classification loss, the object alignment loss and the patch alignment loss with $L=L^{bbox} + L^{cls} + L^{OA} + L^{PA}$.

\begin{table}[t!]\RawFloats
\centering
\caption{\textbf{Architecture Design Ablation}: Comparison between \modelname and ablated versions missing either learnable $det$ tokens (w/o det) or $cls$ feature $o^l_{cls}$ (w/o cls). }
\begin{adjustbox}{width=0.9\textwidth}
\begin{tabular}{l|ccc|ccc|cc}
    \midrule
     \multicolumn{1}{l|}{Model Variant} & \multicolumn{3}{c|}{Refcoco} & \multicolumn{3}{c|}{Refcoco+} & \multicolumn{2}{c}{Refcocog} \\
    & Val & TestA & TestB & Val & TestA & TestB & Val & Test \\
    \midrule
    YORO w/o det. & 74.9 {\color{red} (-8.0)} & 79.4 {\color{red} (-6.2)} & 68.8 {\color{red} (-8.6)} & 66.6 {\color{red} (-6.9)} & 73.2 {\color{red} (-5.4)} & 59.2 {\color{red} (-5.7)} & 69.7 {\color{red} (-3.7)} & 68.9 {\color{red} (-5.4)}  \\
    YORO w/o cls.  & 82.4 {\color{red} (-0.5)} & 84.8 {\color{red} (-0.8)} & 76.7 {\color{red} (-0.7)} & 72.8 {\color{red} (-0.7)} & 77.7 {\color{red} (-0.9)} & 64.0 {\color{red} (-0.9)} & 72.4 {\color{red} (-1.0)} & 73.9 {\color{red} (-0.4)}  \\
    YORO & \textbf{82.9} & \textbf{85.6} & \textbf{77.4} & \textbf{73.5} & \textbf{78.6} & \textbf{64.9} & \textbf{73.4} & \textbf{74.3} \\
    \midrule
\end{tabular}
\end{adjustbox}
 \label{tab:arch_compare}
\end{table}

\begin{figure}[t!]\RawFloats
\begin{tabular}{cc}
\includegraphics[width=0.48\linewidth]{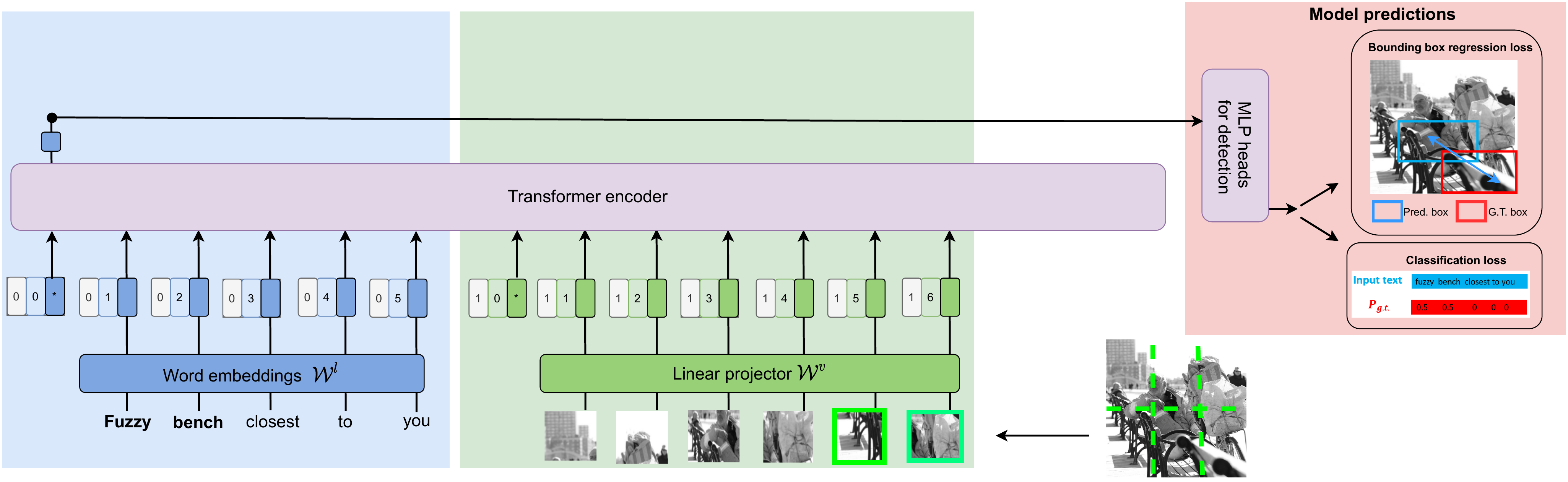}  
& 
\includegraphics[width=0.48\linewidth]{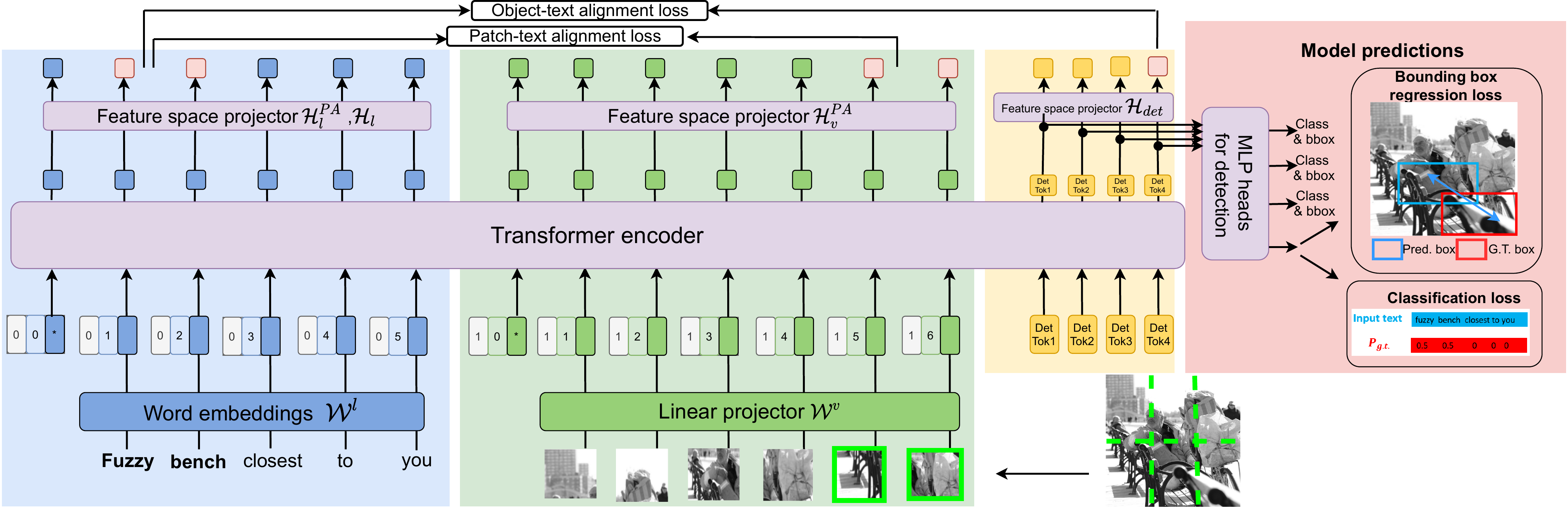}  
 \\
\scriptsize{(a)} & \scriptsize{(b)}
\end{tabular}
\caption{(a) YORO w/o det. token. (b) YORO w/o cls. feature $o^l_{cls}$. }
\label{fig:arch_compare}
\end{figure}

\vspace{-5pt}
\section{Experiments}\label{sec:experiments}\vspace{-5pt}

\subsection{Dataset}\vspace{-5pt}




\modelname is evaluated on five \tasknameshort datasets using the accuracy metric, where the predicted box is correct when the IOU with the ground truth box is at least 0.5. 

\noindent\textbf{RefCoco/RefCoco+/RefCocog}~\cite{ref,Yu2016ModelingCI} are curated from MSCOCO~\cite{Lin2014MicrosoftCC}. RefCOCO consists of 142,209 referred expressions for 50,000 objects in 19,994 images. We adopt the split of~\cite{Deng_2018_CVPR,Zhang_2018_CVPR,Wang_2019_CVPR,ERNIEViL,Liu_2017_ICCV,uniter,vlbert} into train, val, testA, and testB datasets with 120,624, 10,834, 5,657 and 5,095 expressions respectively. 
RefCOCO+ contains 141,564 expressions for 49,856 objects in 19,992 images. We use the split of~\cite{Zhuang_2018_CVPR,Liu_2019_ICCV,ssg,mcn,ERNIEViL,villa}, where train, val, testA and testB datasets contain 120,191, 10,758, 5,726 and 4,889 image text pairs, respectively. 
RefCOCOg 
consists of 85,474 image text pairs of 54,822 objects across 26,711 images. We adopt the UMD split of~\cite{mattnet,Liu_2019_CVPR,Liu_2019_ICCV,Yang_2019_ICCV_DGA,mcn}, into 80,512, 4,896 and 9,602 pairs for train, val and test dataset, respectively.

\noindent\textbf{ReferItGame}~\cite{ref}
contains 19,997 images from the SAIAPR-12~\cite{saiapr} dataset. It has 130,363 text expressions for 99,296 objects. We follow the split of~\cite{cmn, Zhang_2018_CVPR, Luo_2017_CVPR, mattnet, wang2019learning} for training, testing and validation set.

\noindent\textbf{CopsRef}~\cite{copsref}, derived from GQA~\cite{gqa},  contains 508 object classes and average expression length 14.4. When compared to RefCOCO and RefCOCOg (80 object classes and average length around 6), CopsRef has longer expression and more diverse object classes. We adopt the \texttt{WithoutDist} version of the dataset~\cite{copsref,vgtr}, where train, val and test contains 119,628, 12,586 and 16,524 pairs respectively. 

\vspace{-5pt}
\subsection{Training and Evaluation Details}\vspace{-5pt}

We summarize some implementation details of \modelname here to enable reproducibility\footnote{Code available at \url{https://github.com/chihhuiho/yoro}}. \modelname leverages the \texttt{bert-base-uncase}~\cite{bert} pre-trained tokenizer from HuggingFace~\cite{huggingface} with a maximum  text length of 40. The pre-trained checkpoint from \cite{vilt} is used to obtain initial weights. Five detection tokens are used (see Table~\ref{tab:ablade_token_num} for a token number ablation) and randomly initialized using a normal distribution of zero mean and standard deviation 0.02. The AdamW optimizer~\cite{adamw} with initial learning rate of $10^{-4}$ and weight decay of $10^{-2}$ is adopted. This warm up setting is used for the first $10\%$ training epochs and the learning rate is then linearly decayed to 0. \modelname is pre-trained on the concatenated detection dataset curated by \cite{mdetr} for 40 epochs. Since the \tasknameshort task contains a single referred object, image-text pairs with a single bounding box are sampled.  This pre-training allows a good initialization of the detection branch. The pre-trained model is fine-tuned on downstream datasets for 40 epochs. All experiments are conducted using PyTorch~\cite{pytorch} with batch size 128.

\begin{table}[t!]\RawFloats
\begin{minipage}{.65\linewidth}
    \centering
    \caption{Loss Ablation. CL: Classification loss, RE: Bounding box regression loss, OA: Object-text Alignment Loss and PA: Patch-text Alignment loss.}
    \begin{adjustbox}{width=\textwidth}
    \begin{tabular}{cccc|cc|cc}
     \midrule
     \multicolumn{4}{c}{Loss} &  \multicolumn{2}{c}{CopRef} &  \multicolumn{2}{c}{ReferItGame} \\
     CL & RE & OA & PA & Val. & Test & Val. &  Test\\
    \midrule
     \checkmark & \checkmark &  &  & 67.05 & 70.78 & 71.55 & 69.86 \\
     \checkmark & \checkmark & \checkmark &    & 67.73 \textcolor{forestgreen}{(+0.68)} & 71.0 \textcolor{forestgreen}{(+0.22)} & 71.89  \textcolor{forestgreen}{(+0.34)} & 70.39  \textcolor{forestgreen}{(+0.53)}\\
    \checkmark & \checkmark & \checkmark & \checkmark  & \textbf{68.08} \textcolor{forestgreen}{(+1.03)} & \textbf{71.3} \textcolor{forestgreen}{(+0.52)} & \textbf{72.67}\textcolor{forestgreen}{(+1.12)} & \textbf{71.9} \textcolor{forestgreen}{(+2.04)}\\
    \midrule
    \end{tabular}
    \end{adjustbox}
    \label{tab:ablade_loss_cops_ref}
\end{minipage}\quad
\begin{minipage}{.28\linewidth}
    \centering
    \caption{Ablation of number of detection tokens on RefCoco+. } 
    \begin{adjustbox}{width=0.98\textwidth}
    \begin{tabular}{c|ccc}
    \midrule
     \multirow{2}{*}{Det. Tok. \#} &  \multicolumn{3}{c}{RefCoco+} \\
      & Val. & TestA & TestB \\
    \midrule
    1 &  73.2 & \textbf{79.1} & \textbf{64.9} \\
    5  & \textbf{73.5} & 78.6 & \textbf{64.9} \\
    10   & 72.8 & 78.78 &   62.8\\
    \hline
    \end{tabular}
    \end{adjustbox}
    \label{tab:ablade_token_num}
\end{minipage}
\end{table}

\vspace{-5pt}
\subsection{Ablation study} \vspace{-5pt}\label{sec:ablation}

\noindent\textbf{Architecture Design.}
The performance of \modelname is compared to the two variants of Fig.~\ref{fig:arch_compare} on RefCoco/RefCoco+/RefCocog, with the results of Table~\ref{tab:arch_compare}. \textit{\textbf{YORO w/o det. token.}} Fig.~\ref{fig:arch_compare}(a) shows a simple extension of the ViLT~\cite{vilt} architecture to VG. In this case, the classification feature of the text modality $o^l_{cls}$ (See Sec.~\ref{sec:trans_output}) is forwarded through the detection head to regress a bounding box, without the use of detection branch or any detection tokens. The first row of Table~\ref{tab:arch_compare} shows that, when compared to YORO, this is between 3.7\% and 8.6\%  worse. For almost all splits, the performance loss is larger than 5\%, demonstrating a clear benefit in using detection tokens for the VG task. \\
\textit{\textbf{YORO w/o cls. feature $o^l_{cls}$.}} Conversely, the use of detection tokens without the holistic classification feature $o^l_{cls}$ is implemented with the architecture of Fig.~\ref{fig:arch_compare}(b). This performs slightly worse (0.4\% to 1\%) than YORO. The holistic feature contributes to YORO's performance but is much less critical than the detection tokens. Altogether, these ablations support the proposed  design.

\vspace{.05in}
\noindent\textbf{Loss Function.}
Table~\ref{tab:ablade_loss_cops_ref} studies the importance of the various losses in Sec.~\ref{sec:losses}. The baseline consists of classification (CL) and  bounding-box regression (RE) losses. Adding object-text (OA) alignment loss improves accuracy by \textcolor{forestgreen}{+0.22\%} and \textcolor{forestgreen}{+0.53\%}.  When the PA loss is optimized, the gain over baseline increases to \textcolor{forestgreen}{+0.52\%} and \textcolor{forestgreen}{+2.04\%} on the test set of CopsRef and ReferItGame respectively. This indicates both OA and PA loss improve detection accuracy on multiple datasets. The improved performance of PA validates the importance of accounting for the inaccuracy of bounding box predictions in the early training epochs.

\vspace{.05in}
\noindent\textbf{Number of Detection Tokens.}
Table~\ref{tab:ablade_token_num} presents an ablation of the number of detection tokens. On average, using 1 or 5 tokens has comparable results, with a slight decrease for 10 tokens. Since the VG task only contains a single referent, the lack of benefit in using a large number of detection tokens is sensible.

\begin{table}[t!]\RawFloats
	\centering
	\caption{\textbf{Refcoco/Refcoco+/Refcocog}. Compared with single-stage models, \modelname achieves the best accuracy. \modelname is one of the smallest models and is also the fastest. }
	\scalebox{0.85}{
		\begin{tabular}{l|c|c|ccc|ccc|cc|c|c}
			\multirow{2}{*}{Method}  & \multicolumn{2}{c|}{Backbone} &  \multicolumn{3}{c|}{Refcoco} & \multicolumn{3}{c|}{Refcoco+} & \multicolumn{2}{c|}{Refcocog} & Param. & \multirow{2}{*}{FPS}\\ 
			& Lang. & Visual  & Val & TestA & TestB & Val & TestA & TestB & Val & Test & (M) & \\
			\midrule
			FAOA~\cite{Yang_2019_ICCV} & Bert & Darknet53  & 72.05  & 74.81 & 67.91 & -  & - & -  & - & - & 182.9 & 18.9\\
			RCCF~\cite{rccf} & BiLSTM   & DLA34  &   -   &  81.06 &  71.85 & - & 70.35 & 56.32 & - & - & - & 40\\
			SSG~\cite{ssg} & BiLSTM   & Darknet53   &   -   &  76.51 &  67.50 & - & 62.14 & 49.27 & 58.80 & - & - & -\\
			MCN~\cite{mcn} & BiGRU   & Darknet53  &   80.08 &  82.29 & 74.98  & 67.16 & 72.86 & 57.31  & 66.46 & 66.01 & - & -\\
			ReSC~\cite{resc} & Bert  & Darknet53  & 76.59 & 78.22 & 73.25 & 63.23 & 66.64 & 55.53 & 64.87 & 64.87 & 179.9 & 15.8\\
			TransVG~\cite{transvg} & Bert  & Res101  & 81.02 & 82.72 & \textbf{78.35} & 64.82 & 70.70 & 56.94 & 68.67 & 67.73 & 149.7 & 18.1\\
			\textbf{\modelname}  & Bert  & Linear  & \textbf{82.9} & \textbf{85.6} & 77.4 & \textbf{73.5} & \textbf{78.6} & \textbf{64.9} & \textbf{73.4} & \textbf{74.3} & \textbf{114.3} & \textbf{41.9}\\
			\midrule
		\end{tabular}
	}
	\label{tab:ref_reuslt}
\end{table}

\vspace{-5pt}
\subsection{Quantitative Comparison}\vspace{-5pt}

\begin{table}[t!]\RawFloats
\begin{minipage}{.48\linewidth}
\centering
 \caption{Compared to single/two-stage methods, YORO achieves SOTA on \textbf{ReferItGame}~\cite{ref}.  }
\begin{adjustbox}{width=\textwidth}
    \begin{tabular}{lc||lc}
    \midrule
     Two-stage  & Acc. & Single-stage(*) & Acc.\\
    \midrule
    CMN~\cite{cmn}            & 28.33 & RCCF\text{*}~\cite{rccf}  & 63.79\\
    VC~\cite{Zhang_2018_CVPR}  & 31.13 & SSG\text{*}~\cite{ssg}   & 54.24\\
    Luo~\cite{Luo_2017_CVPR}  & 31.85 & ReSC-B\text{*}~\cite{resc}   & 64.33\\
    MAttNet~\cite{mattnet}    & 29.04 &  ReSC-L\text{*}~\cite{resc}   & 64.60\\
    Sim. Net~\cite{wang2019learning}   & 34.54 & FAOA\text{*}~\cite{Yang_2019_ICCV} & 59.30\\
    CITE~\cite{cite}         & 35.07 & ZSGNet\text{*}~\cite{ZSGNet}  & 58.63\\
    PIRC~\cite{pirc}           & 59.13 & TransVG\text{*}~\cite{transvg}    & 70.73\\
    DDPN~\cite{Yu2018RethinkingDA}    & 63.00 &
    \textbf{\modelname}\text{*} & \textbf{71.90} \\
    \midrule
    \end{tabular}
    \end{adjustbox}
    \label{tab:referitgame_reuslt}
\end{minipage}\quad
\begin{minipage}{.48\linewidth}
    \centering
     \caption{YORO achieves SOTA on \textbf{CopsRef}~\cite{copsref} w/o using g.t. box. }
    \begin{adjustbox}{width=0.94\textwidth}
    \begin{tabular}{l|c|c}
    \midrule
    Method (\text{*} indicates single stage)    & Proposal & Acc.\\
    \midrule
    GroundeR~\cite{groundr}           &  g.t.  &  75.7\\
    Shuffle-GroundeR   &  g.t.  &   58.5\\
    Obj-Attr-GroundeR  &  g.t.  &  68.8 \\
    MattNet~\cite{mattnet}            &  g.t.  &   77.9 \\
    CM-Att-Erase~\cite{Liu_2019_CVPR}      &  g.t.  &   80.4\\
    MattNet-Mine~\cite{copsref}       &   g.t.  &   78.4\\

    \midrule
    VGTR-ResNet50~\cite{vgtr} & Pred. & 66.73 \\
    VGTR-ResNet101~\cite{vgtr} & Pred. & 67.75 \\    
    ReSC-B\text{*}~\cite{resc} & Pred. & 64.49 \\
    ReSC-L\text{*}~\cite{resc} & Pred. & 65.32 \\
    \textbf{\modelname}\text{*}  &  Pred.  &   \textbf{71.3} \\
    \midrule
    \end{tabular}
    \end{adjustbox}
   
    \label{tab:copsref_reuslt}
\end{minipage}
\end{table}

\begin{figure}[t]\RawFloats
\begin{minipage}{.54\linewidth}
\centering
\includegraphics[width=0.95\linewidth]{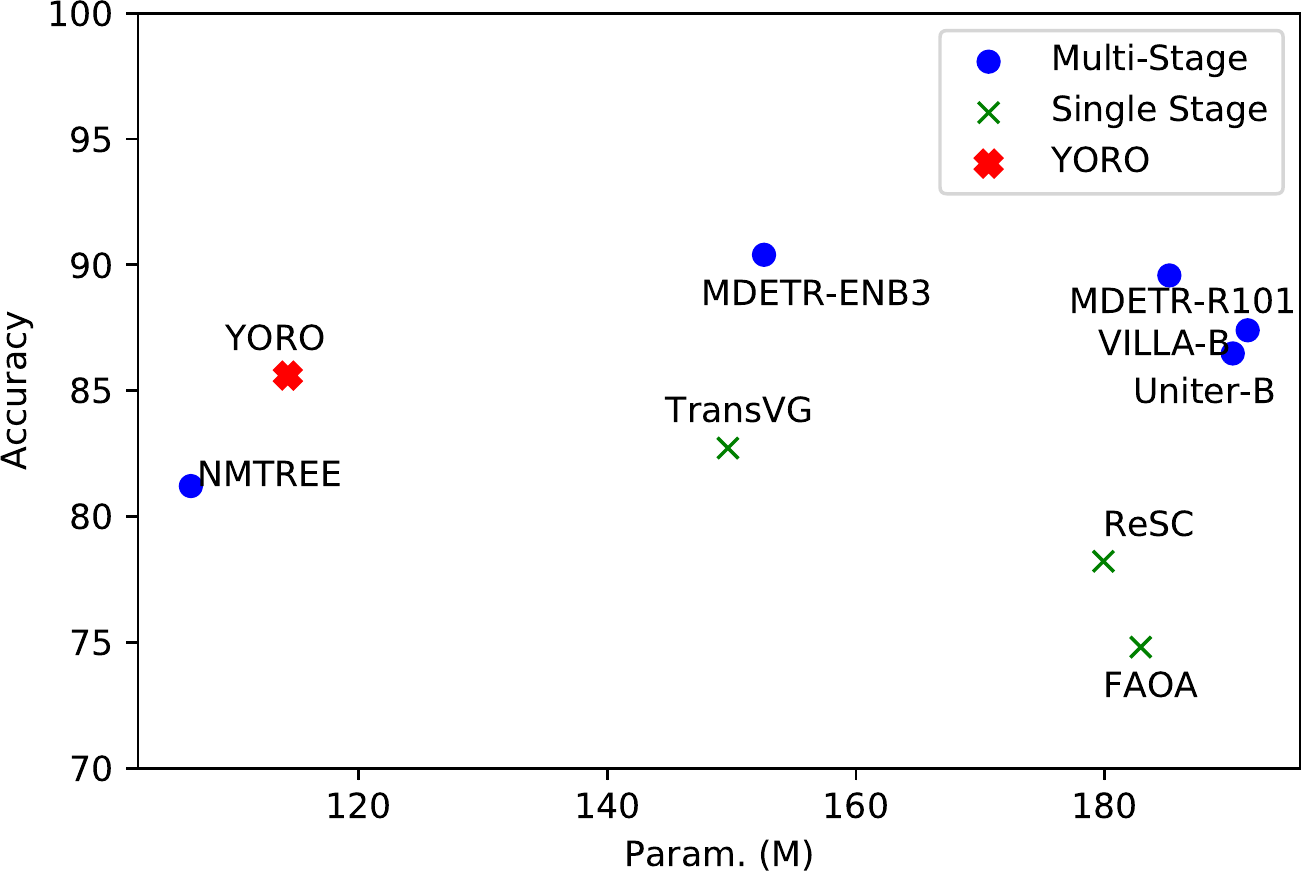}\par
\caption{\modelname beats the prior Pareto front for the memory cost/accuracy trade-off.}
\label{fig:param_acc_plot}
\end{minipage}\hspace{4pt}
\begin{minipage}{.42\linewidth} 
\centering
     \includegraphics[width=0.45\linewidth]{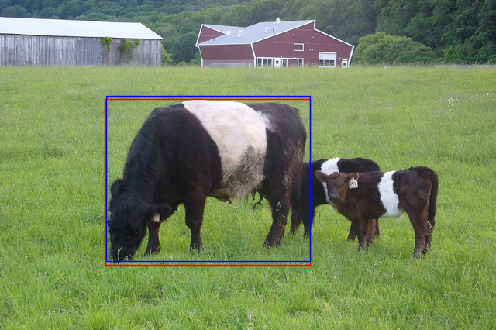} 
     \includegraphics[width=0.45\linewidth]{469.png}
     \\
     \subfloat[][\scriptsize{big cow}]{\includegraphics[width=0.45\linewidth]{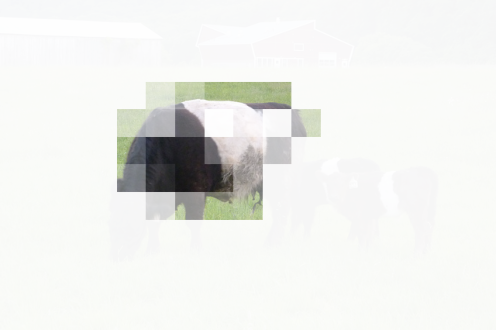}}
     \subfloat[][\scriptsize{blue colored mattress}]{\includegraphics[width=0.45\linewidth]{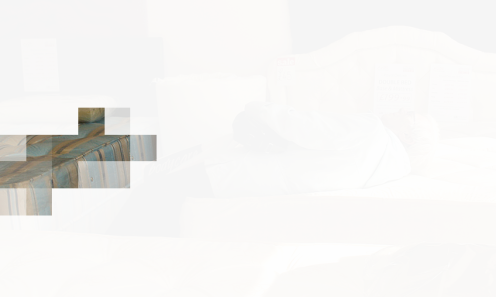}}

     \caption{Top: Predicted (blue) and ground truth (red) box. Bottom: Attention from the selected head.} 
     \label{fig:ref_vis}
\end{minipage}
\end{figure}

\noindent\textbf{RefCoco/RefCoco+/RefCocog.}
Table~\ref{tab:ref_reuslt} summarizes performance of single-stage methods on RefCoco, RefCoco+ and RefCocog for different splits (val, testA and testB).
These are the methods that achieve real time speed (FPS $> 15$) and thus directly comparable to YORO.
\modelname outperforms all single stage methods on seven splits (between abs +1.88\% and +8.6\%). The only exception is Refcoco testB (-0.95\% lower than TransVG~\cite{transvg}). The most competitive method in this class is TransVG, which has $1.3\times$ the size, is $2.3\times$ slower than YORO, and achieves significantly lower accuracies on most splits. The only method of speed comparable to \modelname (RCCF) has significantly lower accuracy on all splits (up to a 14 point drop on Refcocog Val).  A more extensive comparison, including the much heavier two stage models, is presented in the appendix. Overall, as summarized in Fig.~\ref{fig:speed_acc_plot_splash}(b) and Fig.~\ref{fig:param_acc_plot}, \modelname has a significantly better trade-off between speed, parameter size and accuracy than all methods in the literature.

\noindent\textbf{ReferItGame.}
Table~\ref{tab:referitgame_reuslt} compares \modelname to both two-stage (left) and single-stage (right) baselines. \modelname outperforms the best two-stage (DDPN~\cite{Yu2018RethinkingDA}) and single-stage (TransVG~\cite{transvg}) method by 8.9\% and 1.2\% respectively. For roughly the same inference speed, \modelname also beats RCCF~\cite{rccf} by 8.1\%. Note that \modelname achieves SOTA on ReferItGame for both speed and performance.

\noindent\textbf{CopsRef.}
Table~\ref{tab:copsref_reuslt} summarizes the accuracy of \modelname and the baselines on CopsRef~\cite{copsref}. The lower part of the table contains baselines that either detect the referent or select the referent from object proposals. \modelname outperforms the previous SOTA by 3.4\%, without using proposals. The upper part of the table refers to models that use \textit{ground truth} bounding boxes. \modelname fares well even under this unfair comparison, outperforming  some of these baselines. Overall, these results demonstrate that \modelname can generalize to \tasknameshort datasets containing more object classes and longer input text.

\vspace{-5pt}
\subsection{Comparison of Trade-offs Between Size, Speed, and Accuracy }\vspace{-5pt}
To quantify the efficiency of YORO, we compare \modelname with other \tasknameshort baselines  in terms of parameter size and inference speed.

\vspace{.05in}
\noindent\textbf{Inference speed} (FPS) is measured by forwarding 100 images (batch size 1) from the Refcoco testA dataset through the VG model. The results are shown in the rightmost column of Table~\ref{tab:ref_reuslt} and Fig.~\ref{fig:speed_acc_plot_splash}(b). All FPS measurements besides RCCF~\cite{rccf}, MattNet~\cite{mattnet} and DGA~\cite{Yang_2019_ICCV_DGA} were conducted by ourselves, using  a single Titan Xp GPU and  Intel Xeon CPU E5-2630 v4 @ 2.20G\footnote{Note that~\cite{Yang_2019_ICCV} reported a FPS of 26.3 for FAOA, using an NVIDIA 1080TI and Intel Core i9-9900K @ 3.60G set-up, while we measured a FPS of 18.9 under our set-up.}. The FPS of~\cite{rccf,mattnet,Yang_2019_ICCV_DGA} are copied from \cite{rccf}, which measures speed on a Titan Xp GPU (identical to ours) and Intel Xeon CPU E5-2680v4 @ 2.4G (superior to  ours).  \modelname achieves 41.9 FPS and is the fastest \tasknameshort model in the literature.
It is between $3.3\times$ and $14\times$ faster than multi-stage models and at least twice as fast than all single-stage models other than RCCF, which has much weaker accuracy. 

Fig.~\ref{fig:speed_acc_plot_splash}(b) summarizes the trade-off between speed and accuracy of several methods on Refcoco testA. Single-stage methods tend to have significantly lower accuracy than multi-stage methods. The major exception is YORO, which is substantially more accurate than the other single-stage methods, bridging a significant portion of the accuracy gap between the two types of models. Note that \modelname is well above the line connecting MDETR and RCCF, which determines the Pareto front for the \tasknameshort problem. It also has a large gain in inference speed (close to $3.5 \times$ faster) over MDETR. More plots are shown in the supplemental.

We also breakdown the percentage of inference time consumed per YOLO module. Multi-modal VG inputs (Sec.~\ref{sec:trans_input}), transformer encoder (Sec.~\ref{sec:trans_encoder}) and detection head (Sec.~\ref{sec:det_heads}) consume 33.5\%, 62.8\% and 3.7\% of the  inference time, respectively. This suggests that future research should focus on optimizing the design of the transformer layers for both speed and accuracy.

\vspace{.05in}
\noindent\textbf{Parameter sizes} are compared in the 
second column from the right of Table~\ref{tab:ref_reuslt} and
Fig.~\ref{fig:param_acc_plot}. \modelname is smaller than all baselines besides NMTREE~\cite{Liu_2019_ICCV}, which uses a LSTM based language model.
Note that, in particular, \modelname is the smallest of the single-stage methods (See green $\times$ in Fig.~\ref{fig:param_acc_plot}). 

\begin{figure}[t!]\RawFloats
     \centering
     
     \hspace{1pt}
     \subfloat[][\scriptsize{remote at 900} ]{\includegraphics[width=0.23\linewidth]{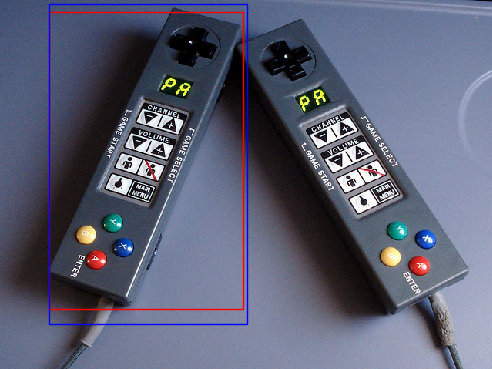}} 
     \subfloat[][\scriptsize{raspberries }]{\includegraphics[width=0.23\linewidth]{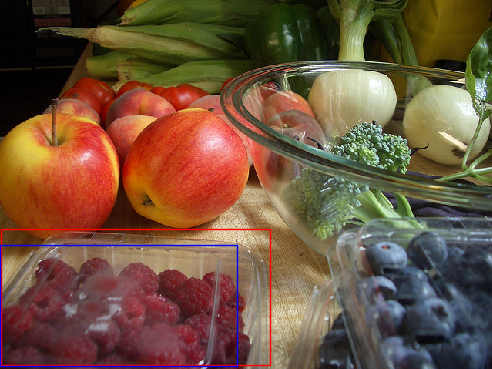}}
     \hspace{1pt}
     \subfloat[][\scriptsize{a sandwich with colby jack cheese , tomato , and lettuce , on fresh cut bread}]{\includegraphics[width=0.23\linewidth]{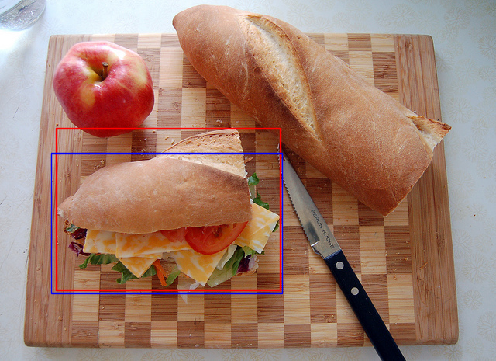}} \hspace{1pt} 
     \subfloat[][\scriptsize{a computer screen with a blue background and itunes open}]{\includegraphics[width=0.23\linewidth]{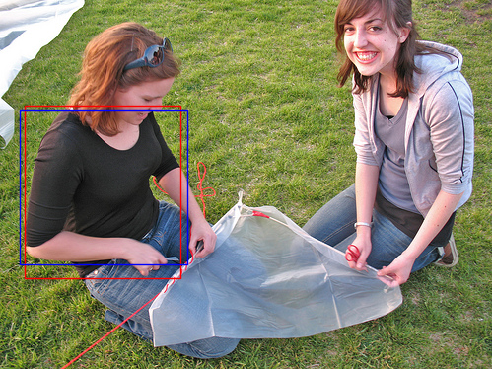}} \\
     \subfloat[][\scriptsize{horse 2nd from left}]{\includegraphics[width=0.23\linewidth]{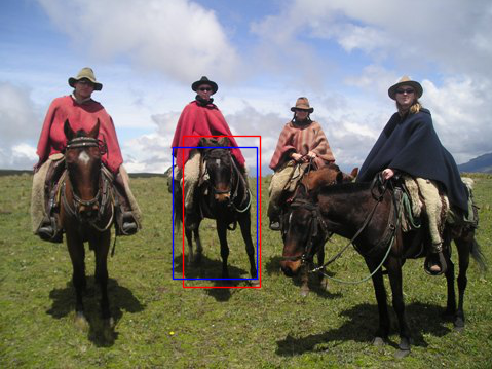}}  \hspace{1pt}
     \subfloat[][\scriptsize{blue bldg , top floor leftest window}]{\includegraphics[width=0.23\linewidth]{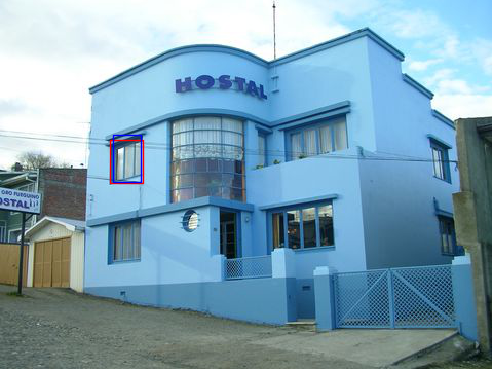}} \hspace{1pt}
      \subfloat[][\scriptsize{The cat that is not brown and that is not to the right of the chair}]{\includegraphics[width=0.23\linewidth]{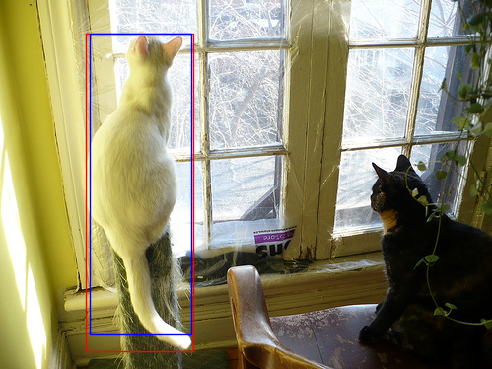}} \hspace{1pt}
     \subfloat[][\scriptsize{The tree that is not leafless and that is not to the right of the sign}]{\includegraphics[width=0.23\linewidth]{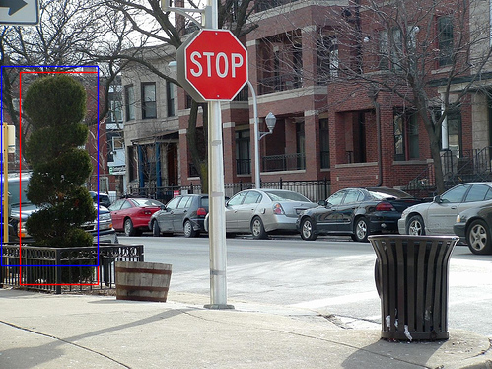}} 
     \caption{\textbf{\modelname Qualitative}: Visualization of the predicted box (\textcolor{blue}{\textbf{blue}}) and ground truth box (\textcolor{red}{\textbf{red}}) in Refcoco+ (a-b), Refcocog (c-d), ReferItGame (e-f) and CopsRef (g-h). The input text for the referent is listed below each image.} 
     \label{fig:vg_vis}
\end{figure}

\vspace{-5pt}
\subsection{Qualitative Results}\vspace{-5pt}
Fig.~\ref{fig:ref_vis} shows the patches of higher attention weight for two example images, computed as follows. Let the detection token corresponding to the predicted box be token $k$. The attention weight between each patch and detection token $k$ is converted to an heatmap and overlaid on the input image. It is clear that YORO attends to the referred objects. See appendix for other examples.
Fig.~\ref{fig:vg_vis} and the appendix show predictions by \modelname on all 5 datasets. \modelname can perform high quality detection not only on short sentences (a-b), but also on convoluted sentences  (c-h). It also performs well in the presence of challenging input text, such as the digits of (a,e), abbreviation of (f), and objects of various sizes (b,f,h).

\vspace{-5pt}
\section{Conclusion} \vspace{-5pt}
In this work, we investigate the speed accuracy trade-off of \tasknameshort models. To pursue a better trade-off, we proposed YORO, a single-stage end-to-end trainable architecture.
A novel patch-text alignment loss is proposed to improve the alignment between language and image features. Experiments show that \modelname outperforms existing single-stage methods. It is the fastest VG and achieves the best trade-off between size, speed, and accuracy in the literature. We believe this is of interest for many resource constrained \tasknameshort applications.

\clearpage

%
%
\bibliographystyle{splncs04}
\bibliography{egbib}

\section*{Appendix - Supplementary material}

\appendix

\section{Limitation and Broader Impact}
In this work, we designed a visual grounding (VG) model which has less parameters than state-of-the-art and has inference speed much faster than real-time (15 fps). \modelname is the fastest VG methods in the literature and out-performs single-stage VG approaches by large margins. It also achieves better speed/accuracy trade-off than multi-stage VG models with $3.3\times$ to $14\times$ faster speed.

We believe our approach being light-weight will facilitate visual grounding (VG) in devices with less computation resources, such as edge devices, mobile robots, low-powered devices (e.g. Raspberry-Pi), and AR/VR devices. In addition, the proposed model requires less computation resources which is likely to be environmental friendly. These aspects warrant further research and consideration.

\section{Patch-Text Supervision}
\begin{figure}[t]
\centering
\begin{tabular}{ccc}
    \includegraphics[width=0.25\linewidth]{epoch0.png} & 
    \includegraphics[width=0.25\linewidth]{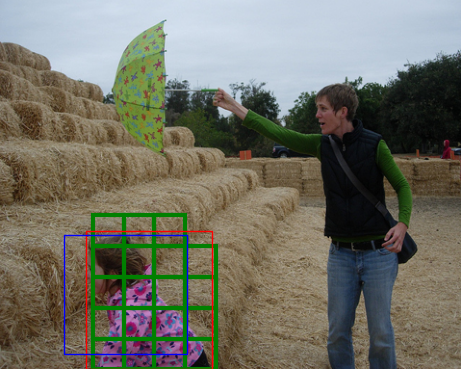} & 
    \includegraphics[width=0.25\linewidth]{epoch5.png} \\
    \scriptsize{Epoch 0} & \scriptsize{Epoch 1} & \scriptsize{Epoch 5}\\
    \includegraphics[width=0.25\linewidth]{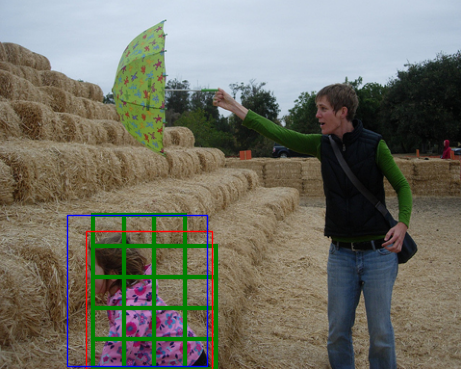} & 
    \includegraphics[width=0.25\linewidth]{epoch20.png} & 
    \includegraphics[width=0.25\linewidth]{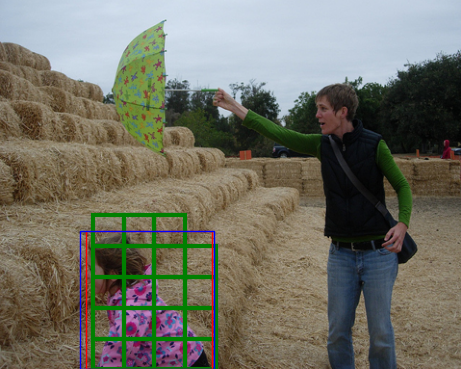} \\
    \scriptsize{Epoch 10} & \scriptsize{Epoch 20} & \scriptsize{Epoch 40}\\
\end{tabular}
\caption{\textbf{Patch-Text Supervision}: Need for consistent supervision for patch-alignment loss. Ground truth box is marked in red and green patches have overlapped with red box with IOU$\geq 0.5$. Blue box is the prediction across different epochs. }
\label{fig:pa_loss_vis_more_epoch}
\vspace{-10pt}
\end{figure}

Figure~\ref{fig:pa_loss_vis_more_epoch} shows the predicted bounding box (blue) and ground truth box (red) of a same image across different epochs. It can be observed that the predicted box gradually becomes more accurate from epoch 0 to epoch 40 and shifts around various locations in different epochs. On the contrary, the green ground truth patches, that have IOU$\geq 0.5$ with the ground truth box, stay consistent throughout the training epochs and make the proposed patch-text alignment loss a more stable. This shows the need for proposed novel patch-text loss.

\section{Parameter Size / Accuracy and Speed /Accuracy Plots}
More comparisons on Parameter size vs Accuracy and Inference Speed vs Accuracy are provided here (w.r.t. Sec. 5 in main paper). We hope the community is inspired to consider other factors which make the technology useful (like inference speed and parameters) and not just on only Accuracy and the race to beat state-of-the-art in performance metrics.

\begin{figure*}
\centering
\begin{tabular}{ccc}
    \includegraphics[width=0.3\linewidth]{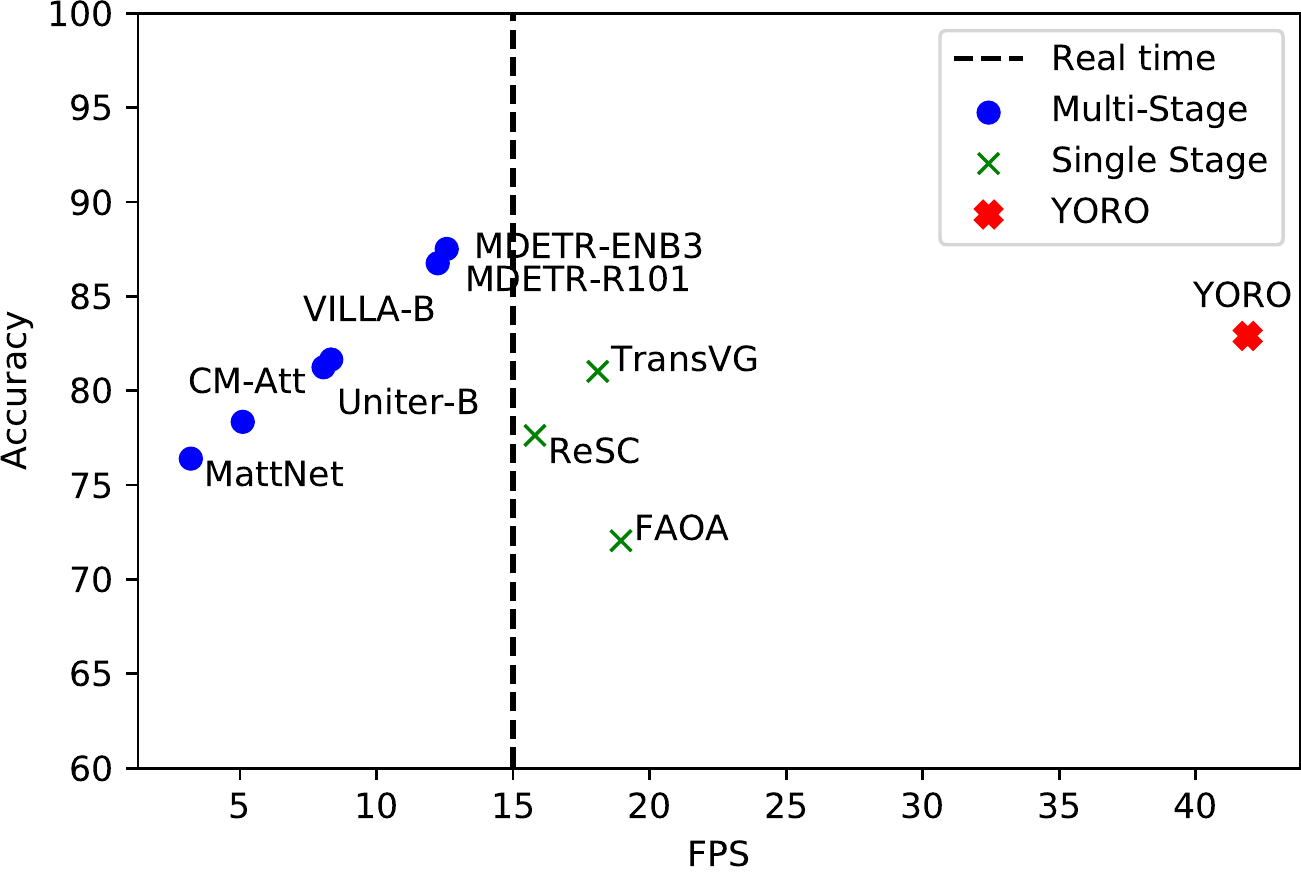} & 
    \includegraphics[width=0.3\linewidth]{refcoco_testA_acc.pdf} & 
    \includegraphics[width=0.3\linewidth]{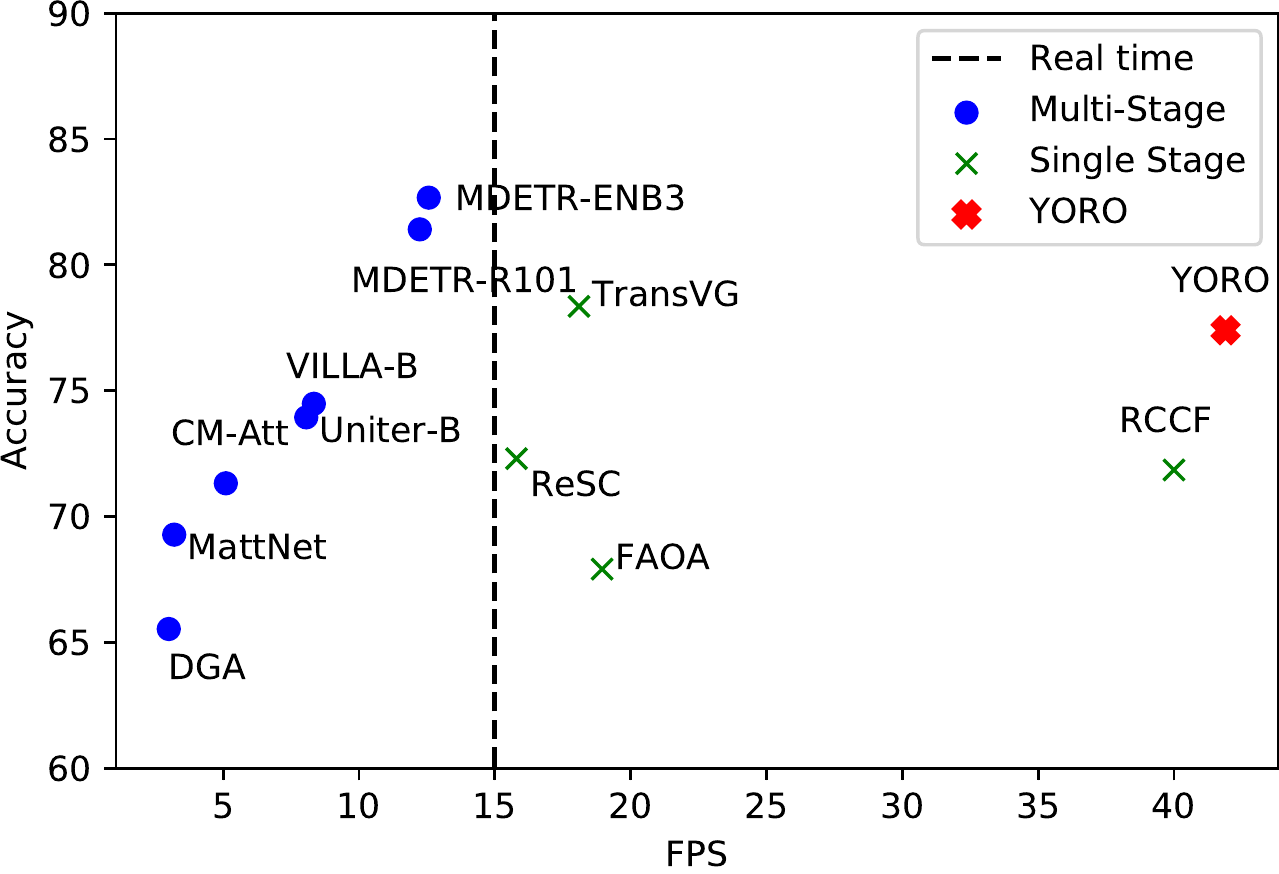} 
\end{tabular}
\caption{Speed Accuracy Plot on RefCoco val, testA and testB dataset.}
\label{fig:speed_acc_refcoco}
\end{figure*}

\begin{figure*}
\centering
\begin{tabular}{ccc}
    \includegraphics[width=0.3\linewidth]{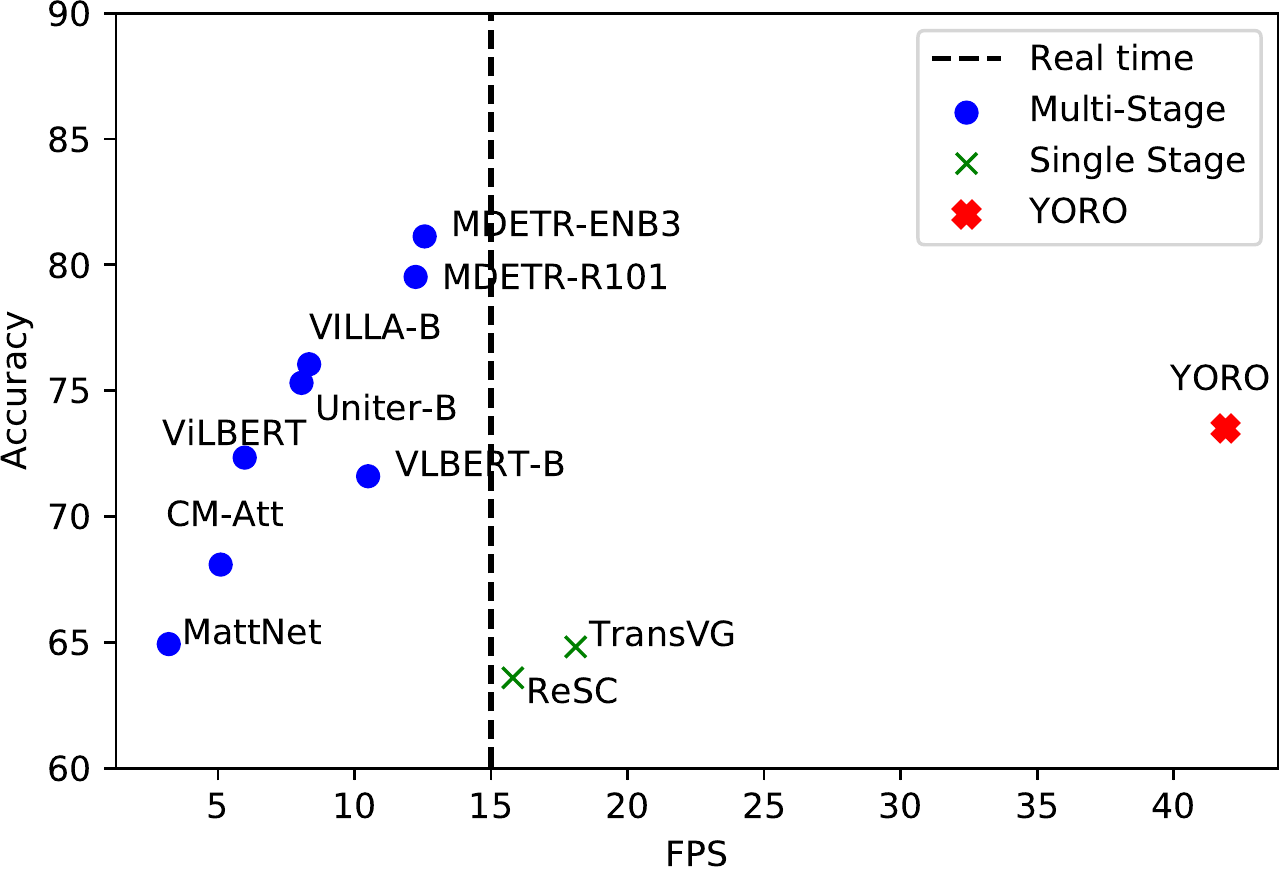} & 
    \includegraphics[width=0.3\linewidth]{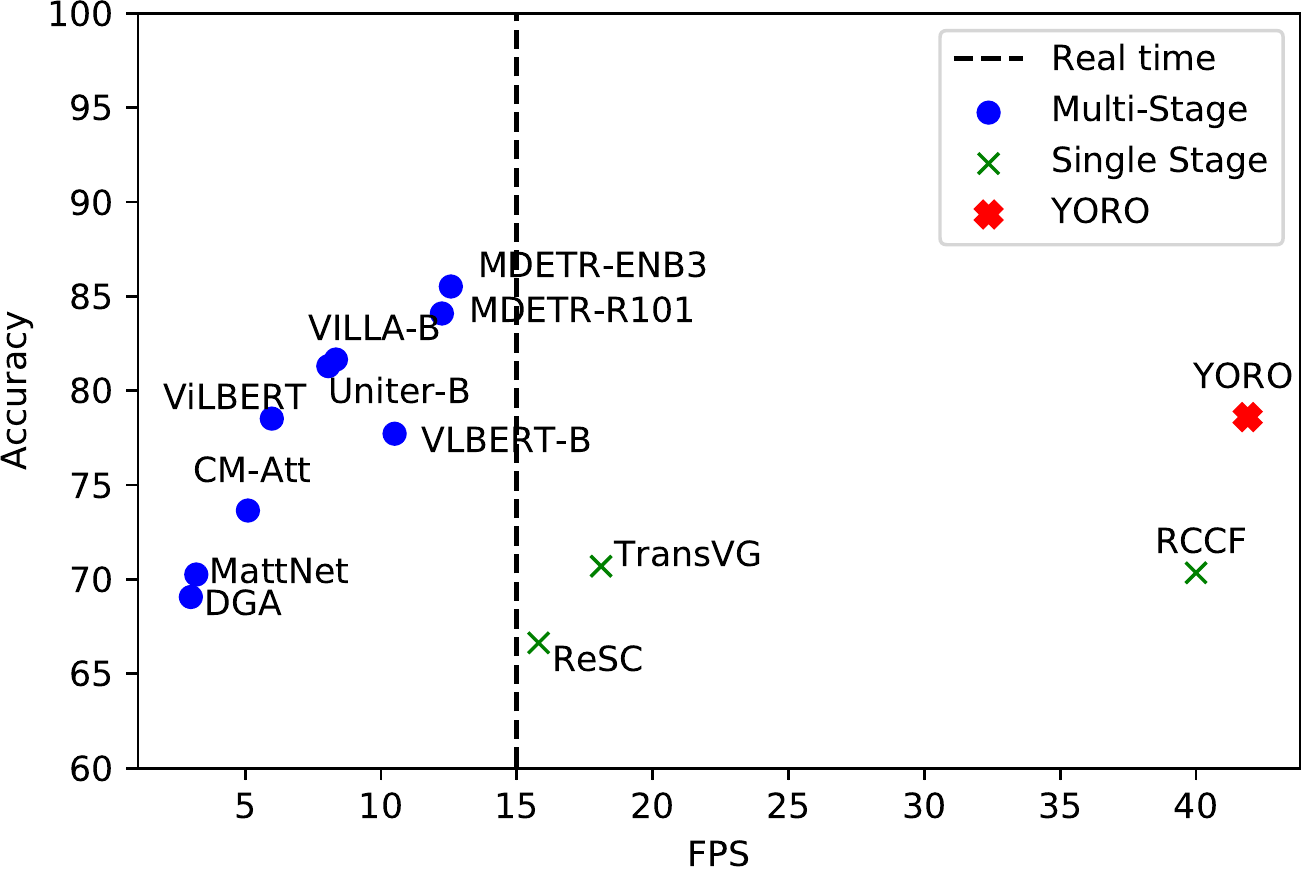} & 
    \includegraphics[width=0.3\linewidth]{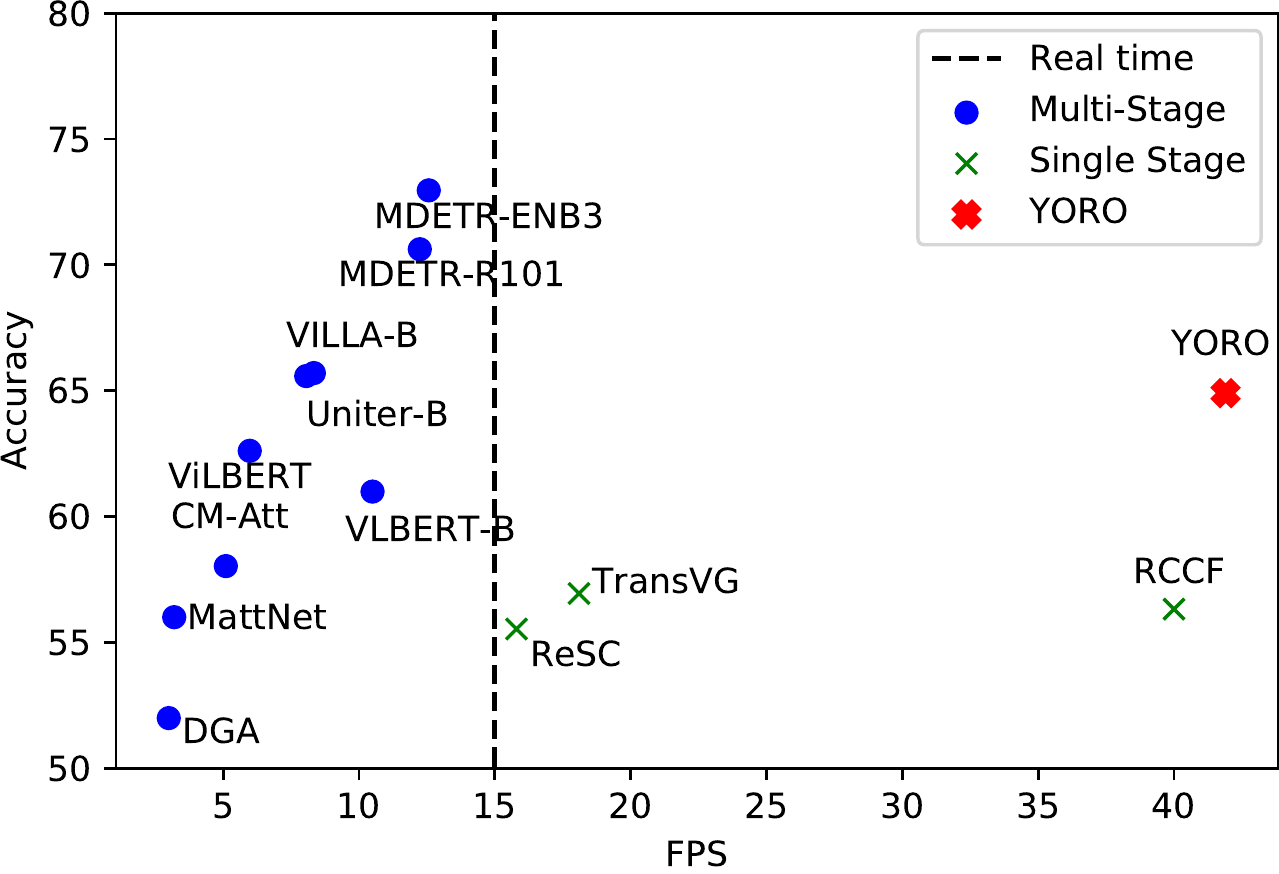} 
\end{tabular}
\caption{Speed Accuracy Plot on RefCoco+ val, testA and testB dataset.}
\label{fig:speed_acc_refcocop}
\end{figure*}

\begin{figure*}
\centering
\begin{tabular}{ccc}
    \includegraphics[width=0.3\linewidth]{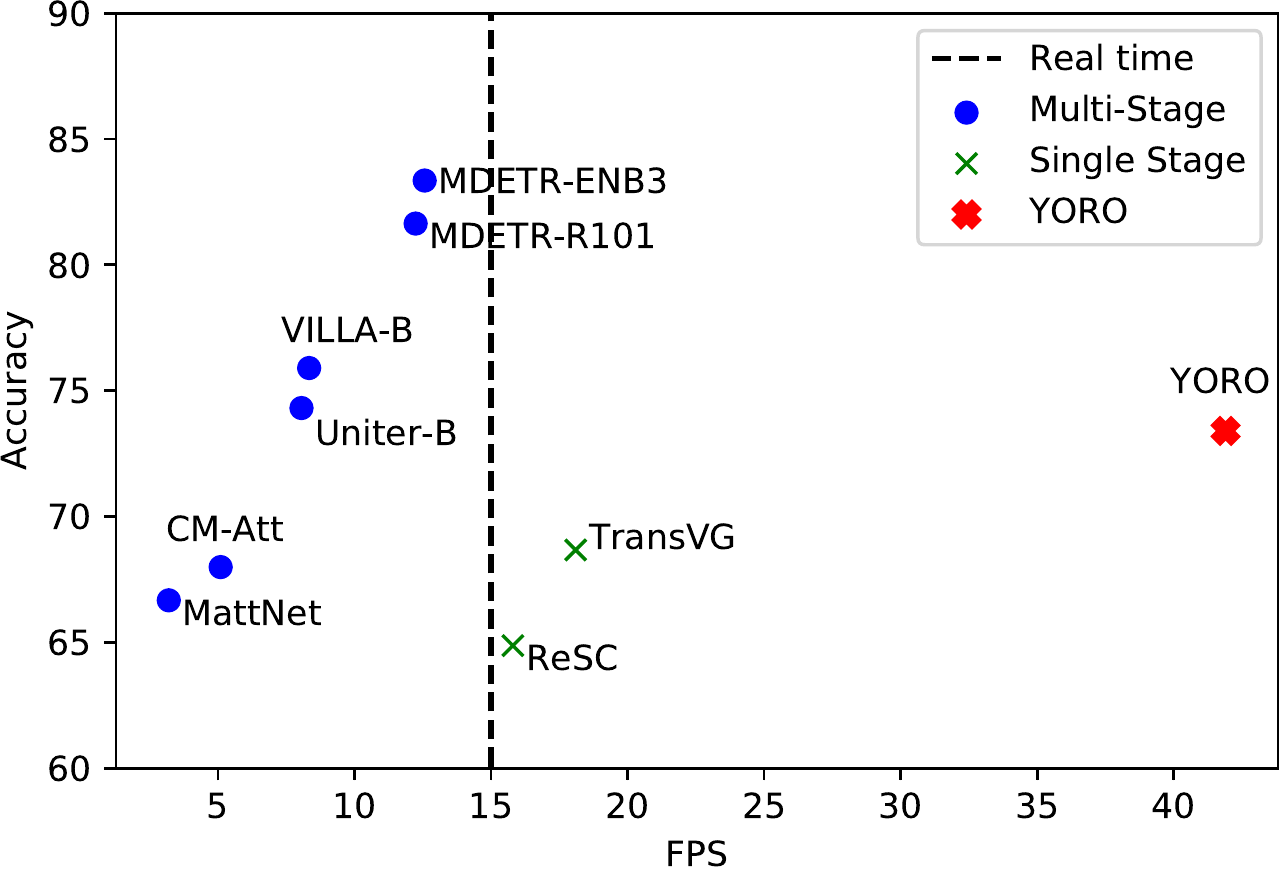} &
    \includegraphics[width=0.3\linewidth]{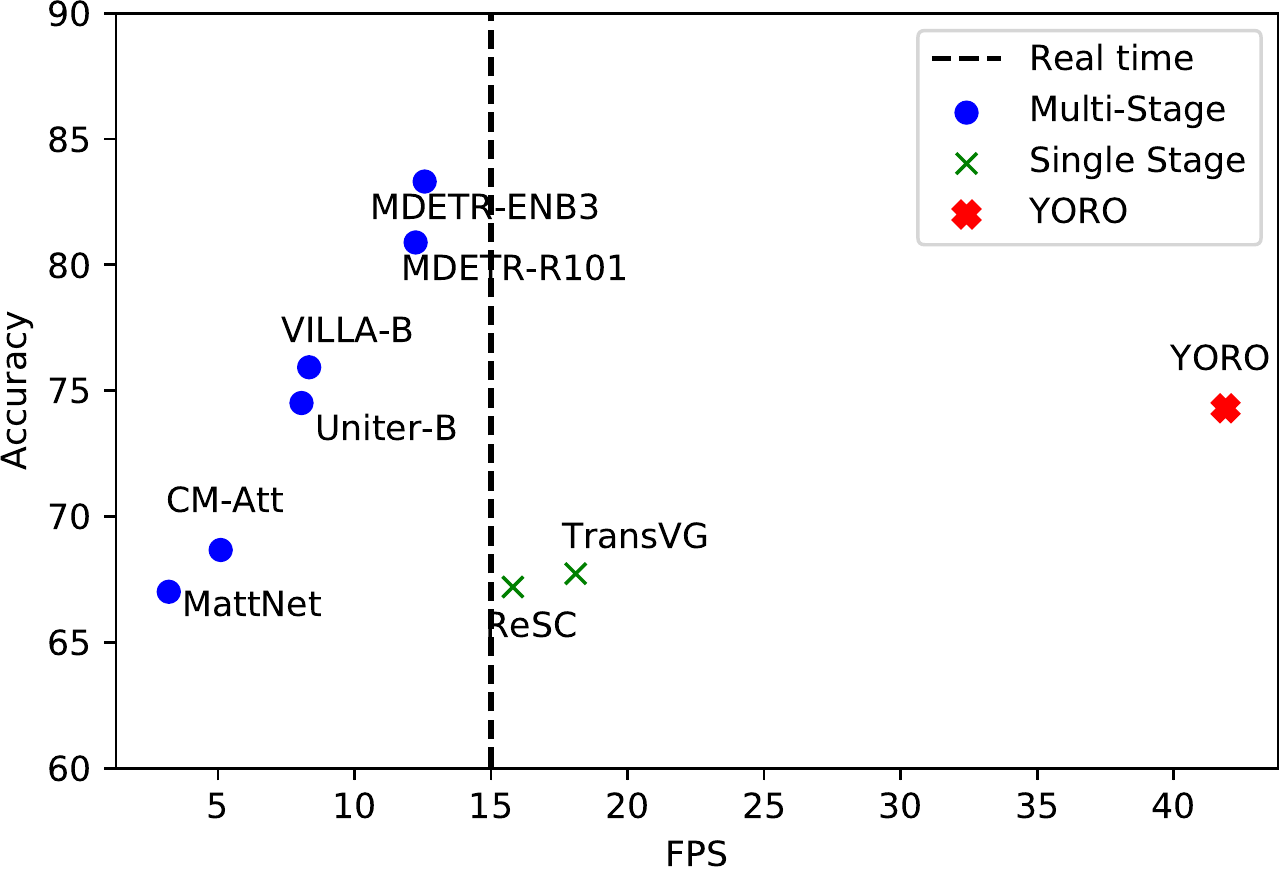}  
\end{tabular}
\caption{Speed Accuracy Plot on RefCocog val, test dataset.}
\label{fig:speed_acc_refcocog}
\end{figure*}

Figure~\ref{fig:speed_acc_refcoco}, \ref{fig:speed_acc_refcocop} and \ref{fig:speed_acc_refcocog} illustrate the trade-off between the accuracy and the inference speed in terms of FPS for RefCoco, RefCoco+ and RefCocog, respectively. YORO achieves better trade-off between the speed and accuracy in different datasets.

\begin{figure*}
\centering
\begin{tabular}{ccc}
    \includegraphics[width=0.3\linewidth]{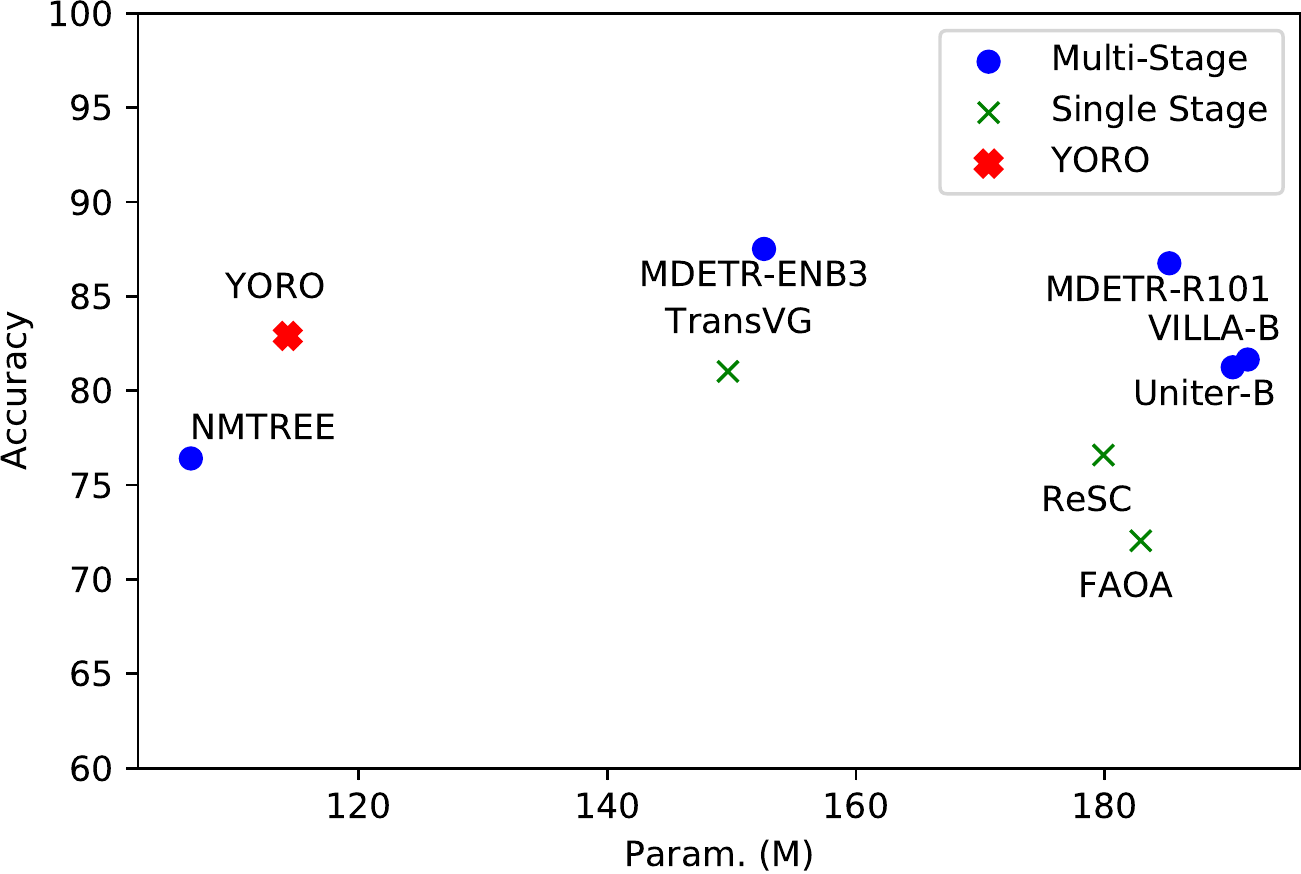} &
    \includegraphics[width=0.3\linewidth]{refcoco_testA_acc_param.pdf} & 
    \includegraphics[width=0.3\linewidth]{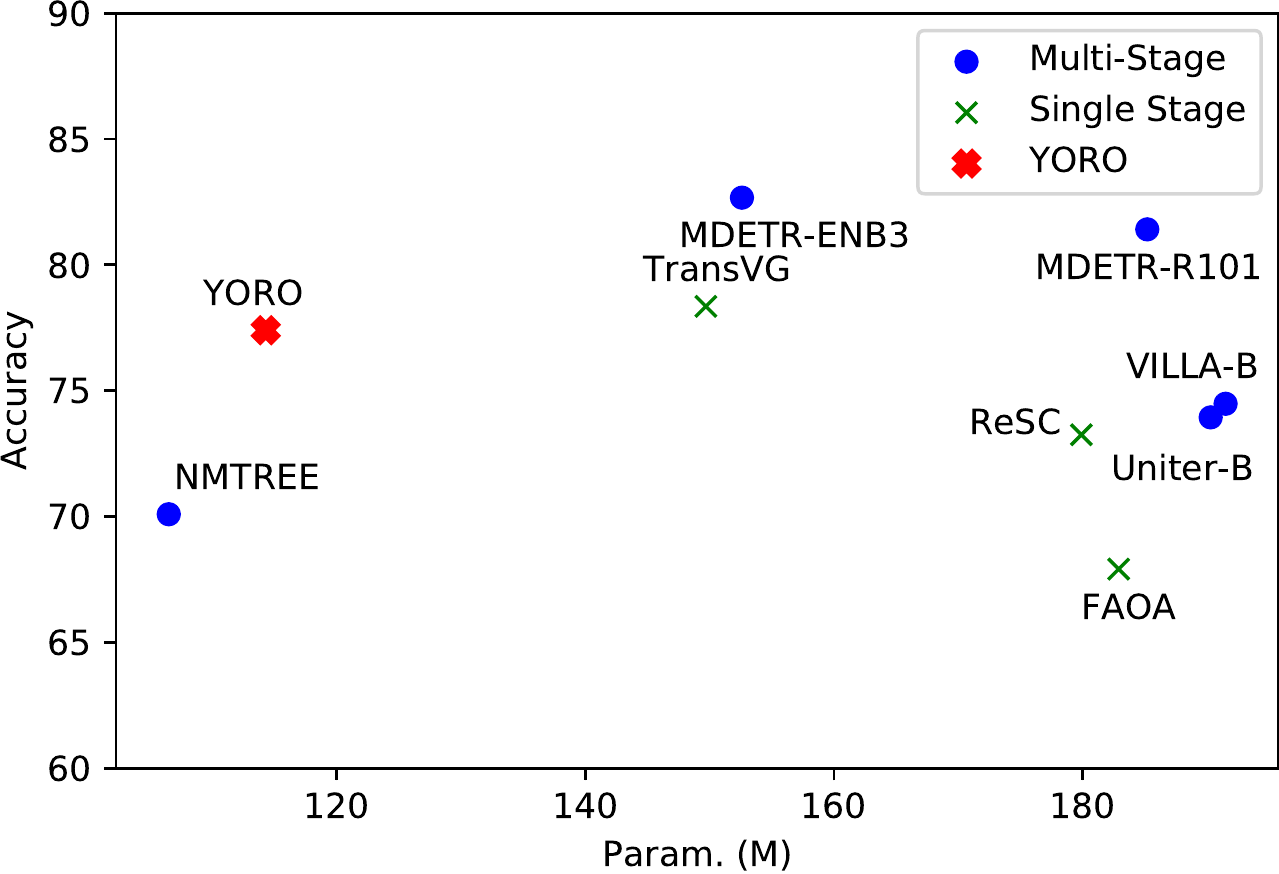}  
\end{tabular}
\caption{Memory Accuracy Plot on RefCoco val, testA and testB dataset}
\label{fig:mem_acc_refcoco}
\end{figure*}

\begin{figure*}
\centering
\begin{tabular}{ccc}
    \includegraphics[width=0.3\linewidth]{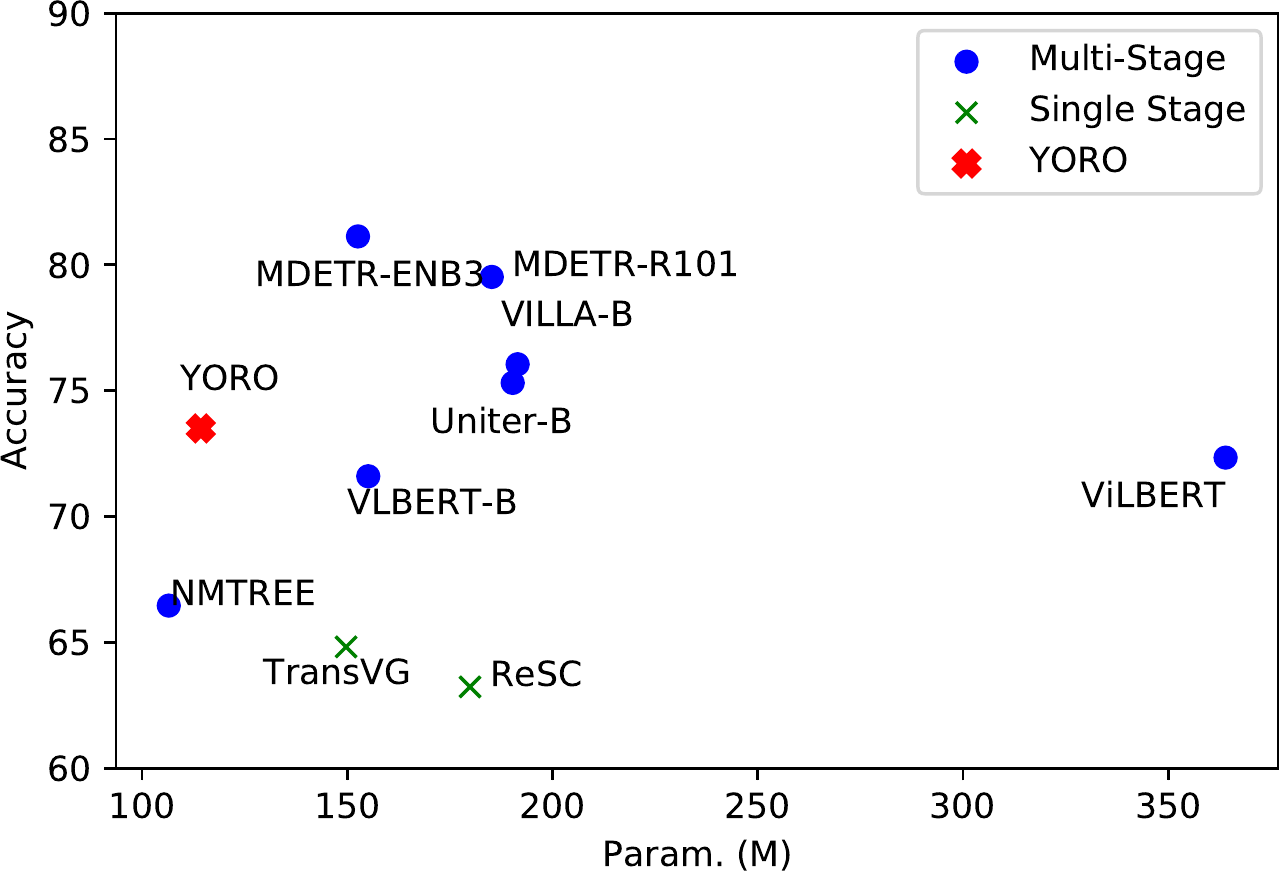} &
    \includegraphics[width=0.3\linewidth]{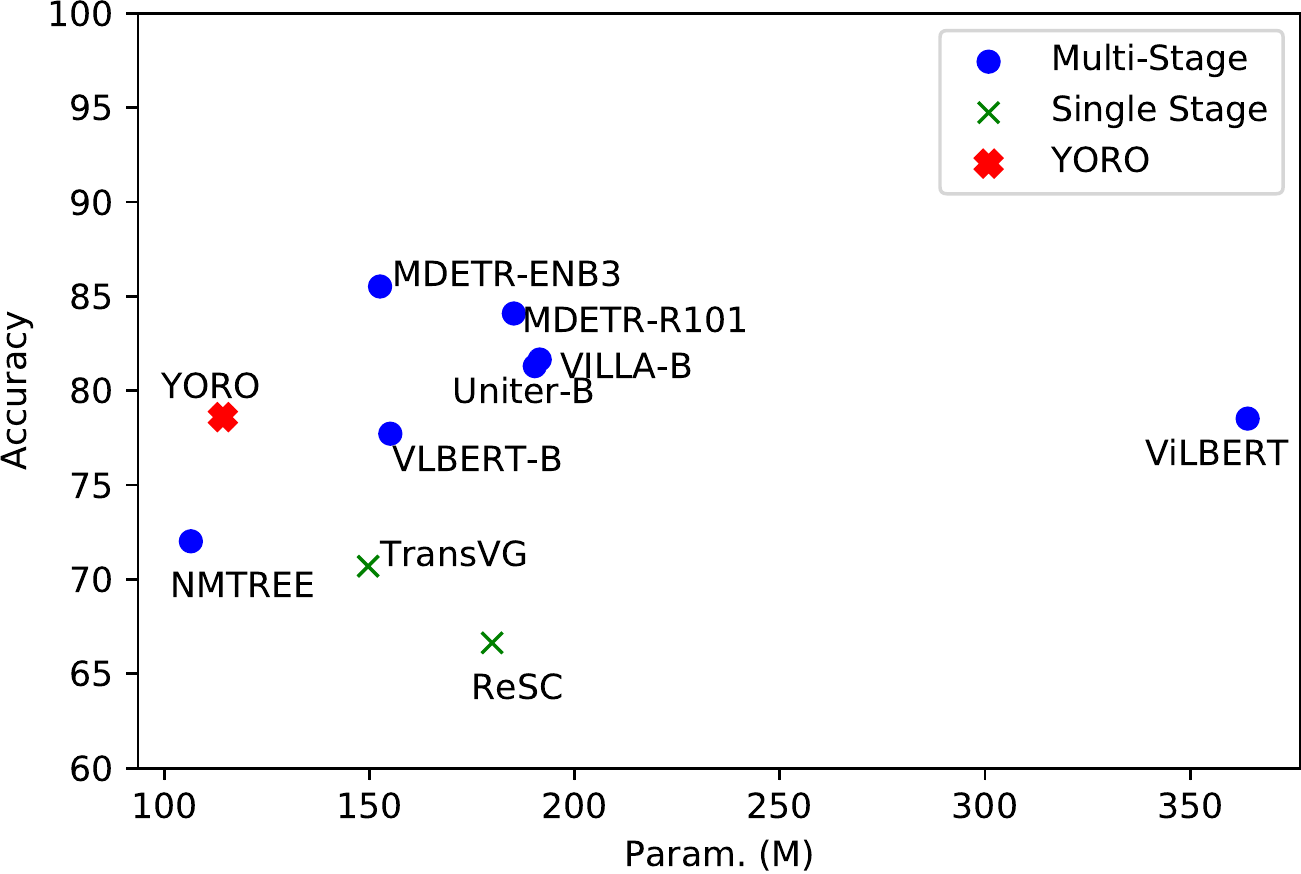} & 
    \includegraphics[width=0.3\linewidth]{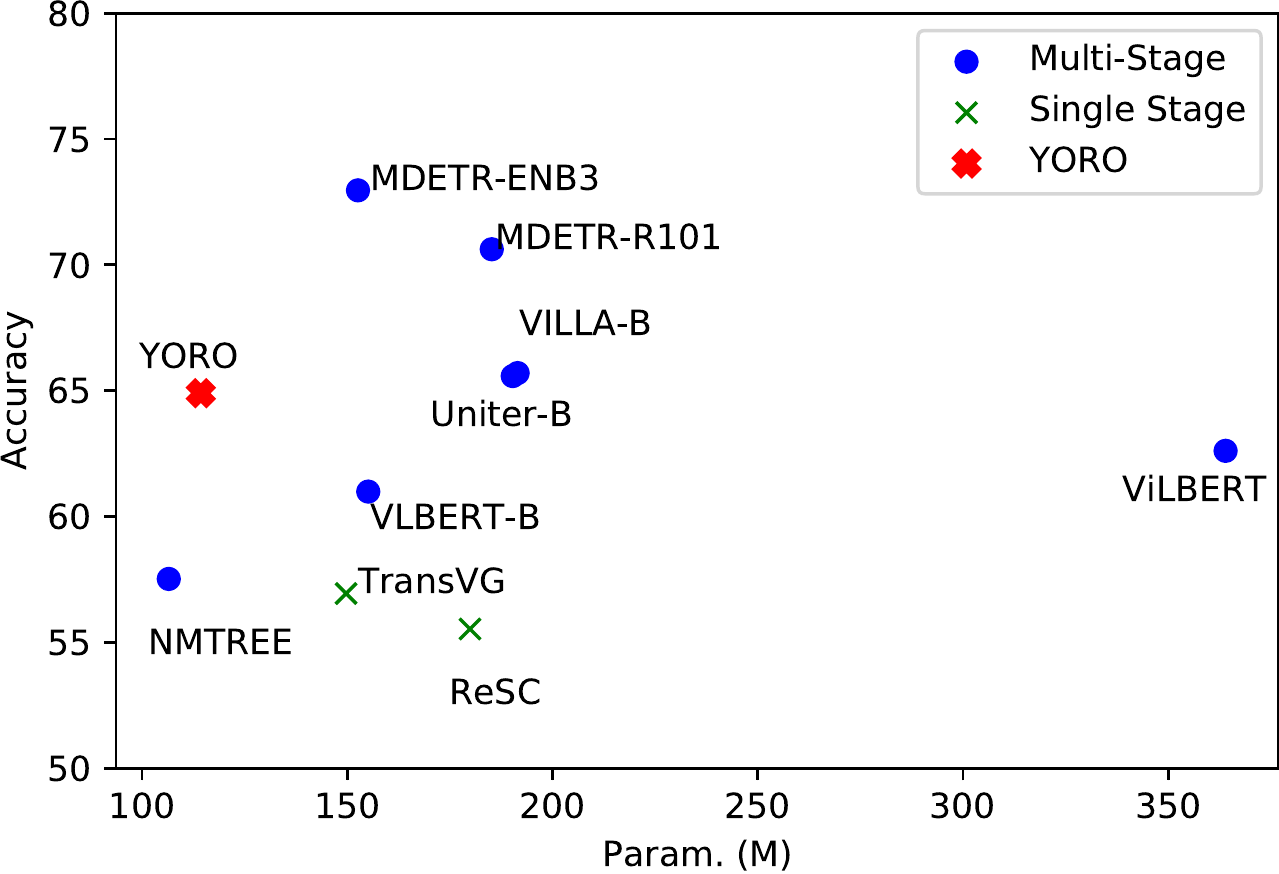}  
\end{tabular}
\caption{Memory Accuracy Plot on RefCoco+ val, testA and testB dataset}
\label{fig:mem_acc_refcocop}
\end{figure*}

\begin{figure*}
\centering
\begin{tabular}{ccc}
    \includegraphics[width=0.3\linewidth]{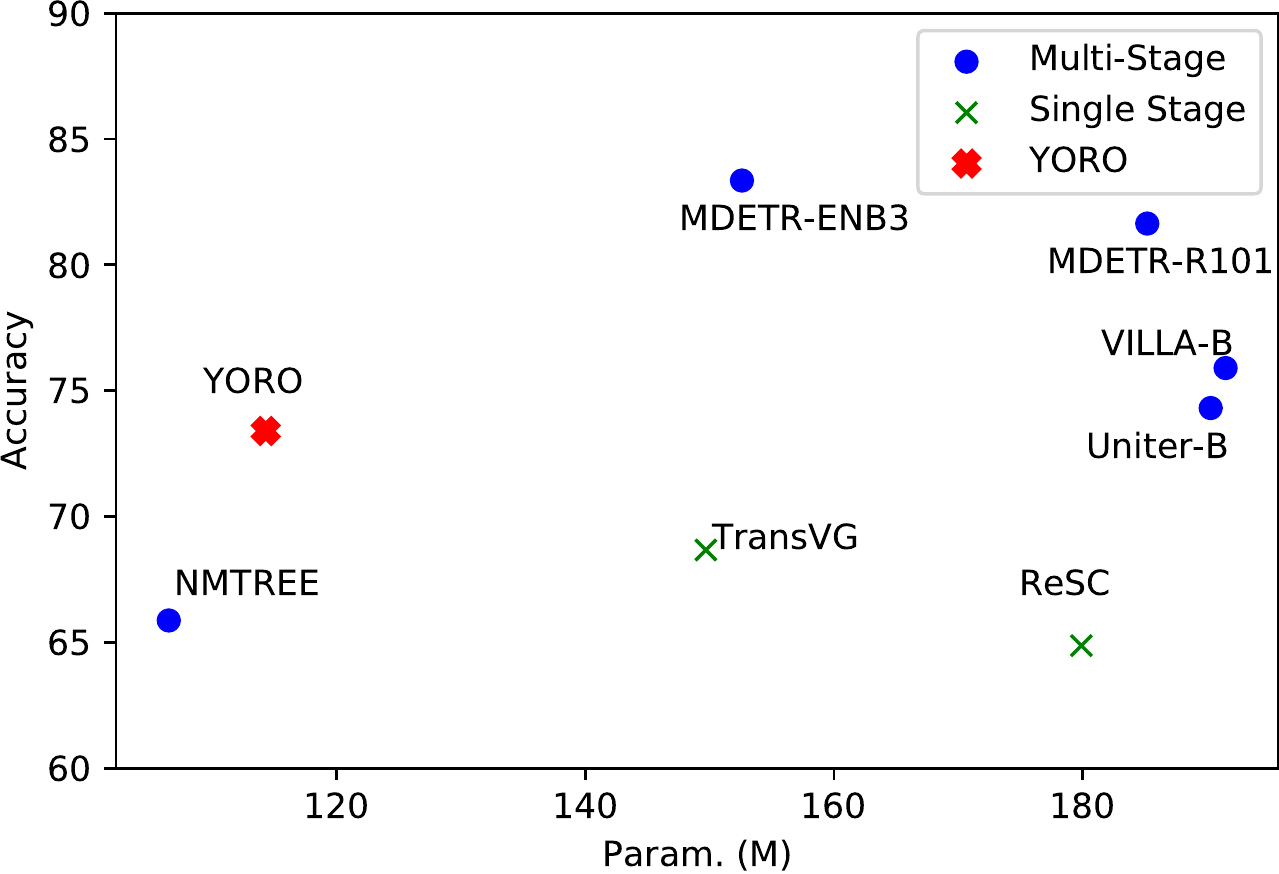} &
    \includegraphics[width=0.3\linewidth]{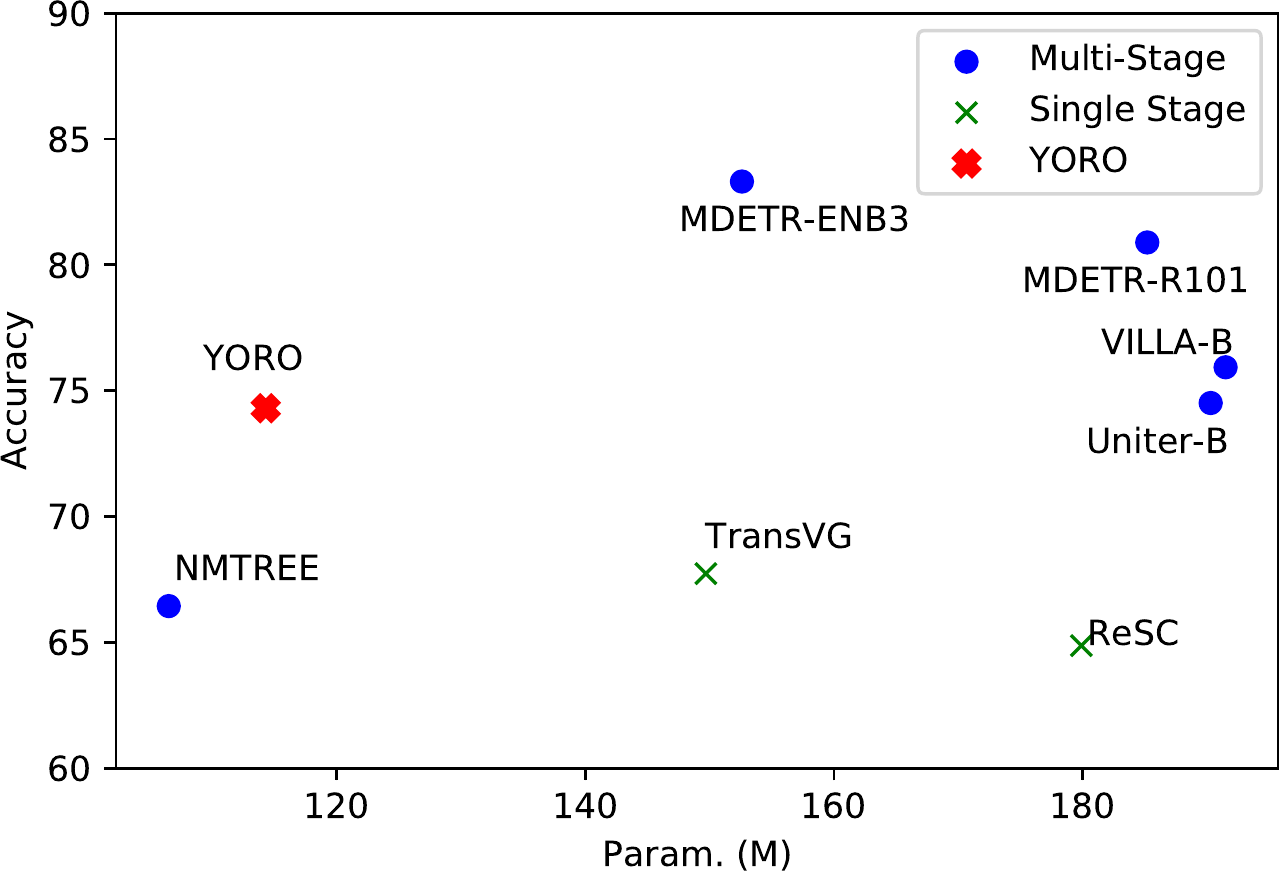}  
\end{tabular}
\caption{Memory Accuracy Plot on RefCocog val, test dataset}
\label{fig:mem_acc_refcocog}
\end{figure*}

Figure~\ref{fig:mem_acc_refcoco}, \ref{fig:mem_acc_refcocop} and \ref{fig:mem_acc_refcocog} illustrate the trade-off between the accuracy and the number of parameter used by each baseline for RefCoco, RefCoco+ and RefCocog, respectively. YORO achieves competitive accuracy with smaller model size. 

This indicates YORO is more suitable for edge devices used in robotic applications and AR/VR devices.

\section{Qualitative Results}
More qualitative visualizations from \modelname are shown here (w.r.t Sec. 5 in main paper). 

\begin{figure*}
     \centering
     \includegraphics[width=0.17\linewidth]{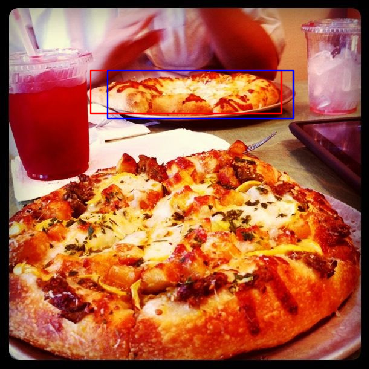}
     \hspace{2pt}
     \includegraphics[width=0.17\linewidth]{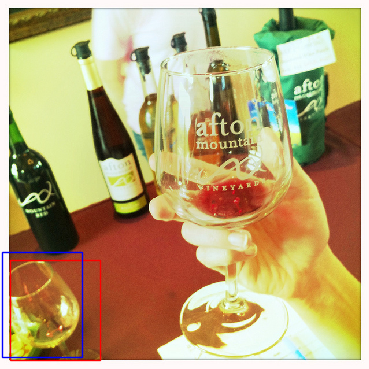}
     \hspace{2pt}
     \includegraphics[width=0.22\linewidth]{394.png}
     \hspace{2pt}
     \includegraphics[width=0.18\linewidth]{142.png}
     \\
     \subfloat[][top pizza]{\includegraphics[width=0.17\linewidth]{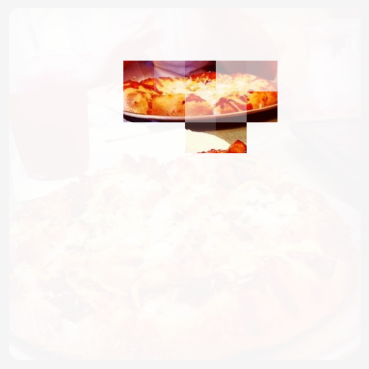}} 
     \hspace{2pt}
     \subfloat[][leftmost small glass]{\includegraphics[width=0.17\linewidth]{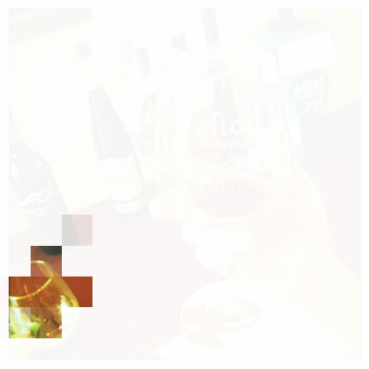}}  
     \hspace{2pt}
     \subfloat[][let green apple]{\includegraphics[width=0.22\linewidth]{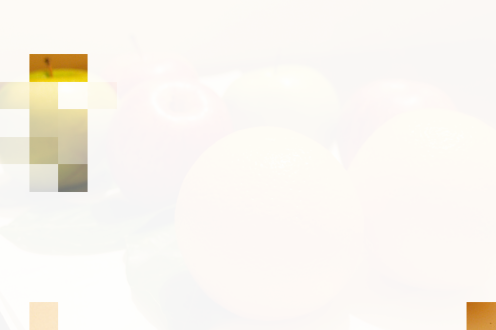}} \hspace{2pt}
     \subfloat[][bottom slice]{\includegraphics[width=0.18\linewidth]{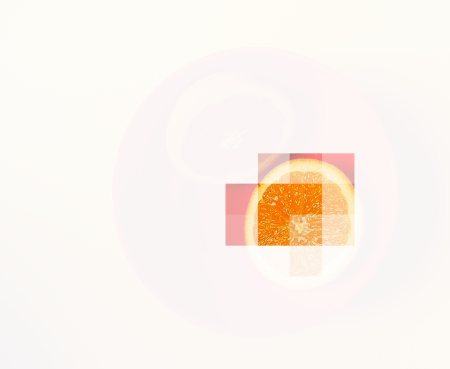}}
     \\
     \includegraphics[width=0.22\linewidth]{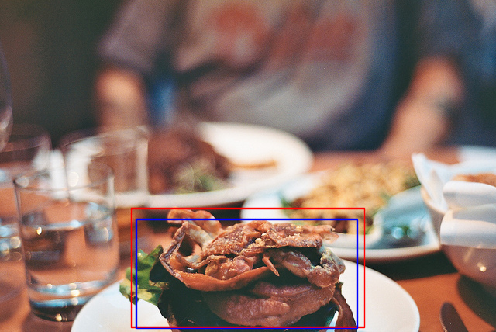}
     \hspace{2pt}
     \includegraphics[width=0.17\linewidth]{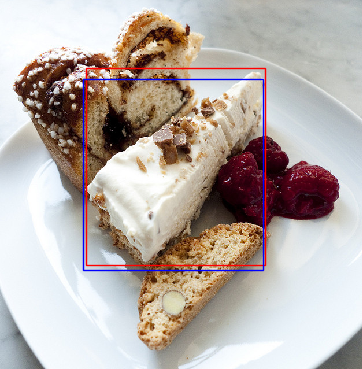}
     \hspace{2pt}
     \includegraphics[width=0.22\linewidth]{466.png}
     \hspace{2pt}
     \includegraphics[width=0.22\linewidth]{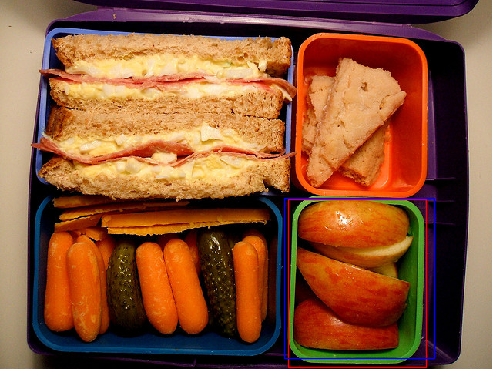}
     \\
     \subfloat[][food nearest to us ]{\includegraphics[width=0.22\linewidth]{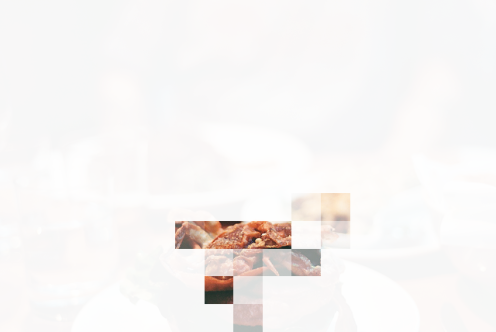}} \hspace{2pt}
     \subfloat[][middle piece ]{\includegraphics[width=0.17\linewidth]{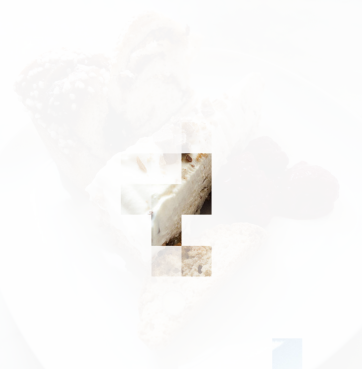}}  \hspace{2pt}
     \subfloat[][clock at 1115]{\includegraphics[width=0.22\linewidth]{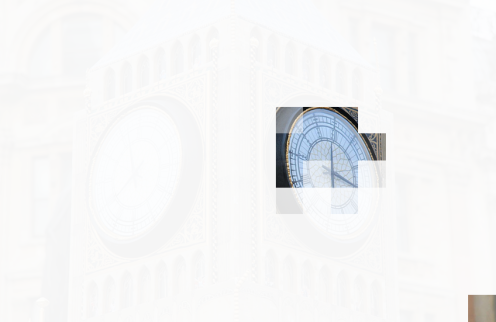}} \hspace{2pt}
     \subfloat[][apples ]{\includegraphics[width=0.22\linewidth]{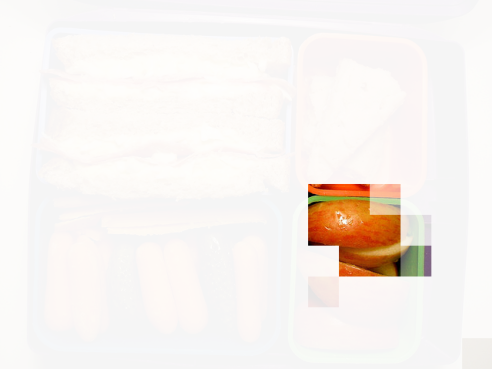}}
     \\
     \includegraphics[width=0.2\linewidth]{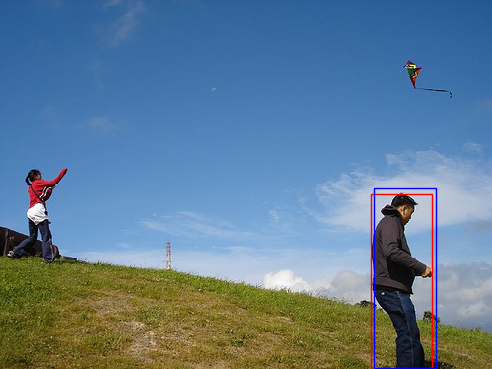}
     \hspace{2pt}
     \includegraphics[width=0.18\linewidth]{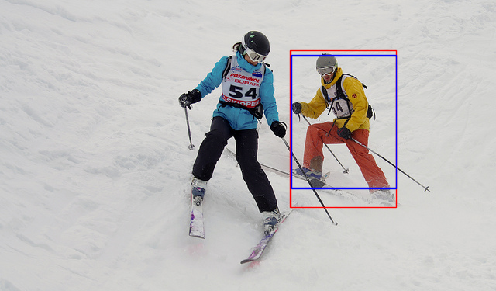}
     \hspace{2pt}
     \includegraphics[width=0.14\linewidth]{24.png}
     \hspace{2pt}
     \includegraphics[width=0.17\linewidth]{41.png}
     \\
     \subfloat[][a man standing next to a young girl on a grassy hillside]{\includegraphics[width=0.20\linewidth]{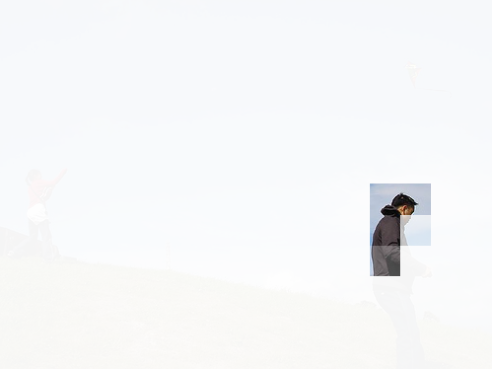}} \hspace{2pt}
     \subfloat[][skiier in red pants]{\includegraphics[width=0.18\linewidth]{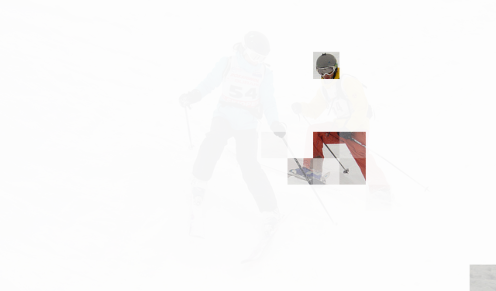}}  \hspace{2pt}
     \subfloat[][a parked white ford suv]{\includegraphics[width=0.14\linewidth]{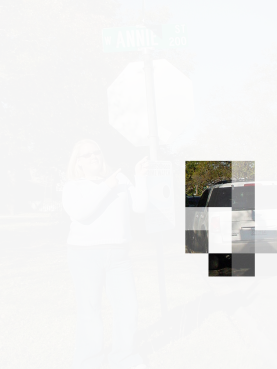}} \hspace{2pt}
     \subfloat[][the orange closest to the banana]{\includegraphics[width=0.17\linewidth]{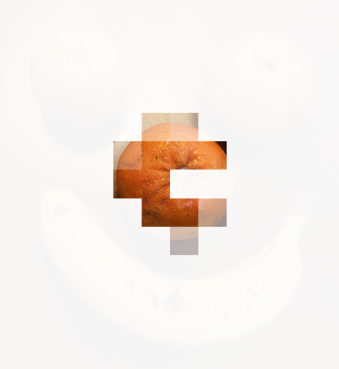}}

     \caption{Visualization of the predicted box (blue) and ground truth box (red) in Refcoco (a-d)/Refcoco+(e-h)/Refcocog(i-l) dataset. The patches that have higher attention are highlighted in the even rows.} 
     \label{fig:ref_vis_appendix}
\end{figure*}

\noindent\textbf{Correct Examples.}
Figure~\ref{fig:ref_vis_appendix} contains correct detection results from YORO output. Examples from RefCoco and RefCoco+ have shorter query text, while examples from RefCocog are associated to more descriptive text. YORO can handle query text of various length and correctly detect the referent when the query contains digits. While the odd rows contain the predicted bounding box, the even rows show the patches that higher attention weight. More specifically, assuming the $k^{th}$ detection token corresponds to the predicted bounding box, we extract the transformer attention weight between each patch and the $k^{th}$ detection token. The attention weight is then converted to heatmap and overlaid on the input image. Take Figure~\ref{fig:ref_vis_appendix}(d) for example. Most patches around the bottom orange have higher attention weight. 

\clearpage
\section{Full Comparison with State of the Art}

Table~\ref{tab:ref_reuslt_appendix} extends the comparison of YORO with both single stage and multi-stage methods. Note that comparing YORO with multi-stage methods in terms of accuracy is \textbf{not an apples-to-apples comparison}, because multi-stage methods tends to sacrifice speed for accuracy improvement with the use of resource consuming models. The table is divided into blocks with the top performance within each block in bold. 

\noindent\textbf{\textit{(Bi)LSTM language model Two-stage methods}}: These methods are based on small (LSTM-based) language models of relatively few parameters that severely under-perform  transformer-based approaches. Note that, for these models, parameter size does not correlate with inference speed, due to recursive LSTM computations. In fact, these are the slowest models considered. \modelname significantly out-performs all these models in most splits (up +7 abs points), for marginally higher memory complexity ($7\%$ increase) and much faster inference ($10\times$ speed up).

\noindent\textbf{\textit{BERT language model Two-stage methods}}: The main difference between \modelname and these methods is the elimination of the visual backbone. 
This drastically reduces parameter size (between $60\%$ and $74\%$ of the parameters) and boosts inference speed by a substantial amount (between $4\times$ and $7\times$ speed-up). Despite these gains, \modelname is comparable to the best of these methods, achieving superior performance on Refcoco val/testB (between +1.25 and +2.9) and comparable in other splits.

\noindent\textbf{\textit{Encoder-Decoder multi-stage methods}}: Like \modelname, these methods are transformer-based but use more powerful encoder-decoder models. While it has been argued that encoder-decoder models have more powerful predictive capability than encoder-only approaches~\cite{t5,detr,yolos}, \modelname outperforms all the methods in this class other than MDETR by large margins (between abs +3.0\% and +10.1\%). The comparison to MDETR is more complex, because it also uses the more powerful RoBERTa~\cite{roberta} language model. This enables MDTER to achieve higher accuracies, but also makes it a lot larger ($1.3 \times$) and slower ($3.4\times$) than \modelname.

\noindent\textbf{\textit{Single-stage methods}}: 
These are the methods directly comparable to \modelname. \modelname outperforms all these methods on seven splits (between abs +1.88\% and +8.6\%). The only exception is Refcoco testB (-0.95\% lower than TransVG~\cite{transvg}). The most competitive method in this class is TransVG, which has $1.3\times$ the size and is $2.3\times$ slower than \modelname but achieves significantly lower accuracies on most of the splits. On the other hand, the only method of speed comparable to \modelname (RCCF) has significantly lower accuracy on all splits (up to a 14 point drop on Refcocog Val). Overall, \modelname has a significantly better trade-off between speed, parameter size and accuracy than all these methods.

\begin{table}[t!]\RawFloats
	\centering
	\caption{\textbf{Refcoco, Refcoco+ and Refcocog}. \modelname achieves SOTA performance when compared with single-stage models. \modelname is one of the smallest models and is also the fastest. \textbf{Bold type} indicates the best model in each block. }
	\scalebox{0.7}{
		\begin{tabular}{l|c|c|ccc|ccc|cc|c|c}
			\multirow{2}{*}{Method}  & \multicolumn{2}{c|}{Backbone} &  \multicolumn{3}{c|}{Refcoco} & \multicolumn{3}{c|}{Refcoco+} & \multicolumn{2}{c|}{Refcocog} & Param. & \multirow{2}{*}{FPS}\\ 
			& Lang. & Visual  & Val & TestA & TestB & Val & TestA & TestB & Val & Test & (M) & \\
			\midrule
			\multicolumn{13}{l}{\textit{\textbf{Two-Stage} Vision-Language methods based on \textbf{(Bi)LSTM} for Language Model}} \\ 
			\midrule
			S.L.R.~\cite{slr} & LSTM & VGG16  & - & 72.94 & 62.98 & - & 58.68 & 47.68 & -  & - & - & -\\  
			Luo~\cite{Luo_2017_CVPR} & BiLSTM & VGG16  & - & 67.94 & 55.18 & - & 57.05 & 43.33 & -  & - & - & -\\  
			Deng~\cite{Deng_2018_CVPR} & LSTM & VGG16 & \textbf{81.27} & 81.17 &  \textbf{80.01}  &  65.56  &   68.76  &  60.63  & - & - & - & - \\
			VC~\cite{Zhang_2018_CVPR} & LSTM & VGG16  & - & 73.33 & 67.44 & - & 58.40 & 53.18 & - & - & - & -\\
			LGRAN~\cite{Wang_2019_CVPR} &  LSTM & VGG16  & -  & 76.6 & 66.4 & -  & 64.0 & 53.4  & - & - & - & -\\
			Liu~\cite{Liu_2017_ICCV} &  LSTM & VGG19  & -  & 72.08 & 57.29 & -  & 57.97 & 46.20  & - & - & - & -\\
			MattNet~\cite{mattnet} & BiLSTM   & Res101   &  76.40 & 80.43 & 69.28 & 64.93 & 70.26  & 56.00 & 66.67 &  67.01  & - & 3.2\\
			CM-Att-Erase~\cite{Liu_2019_CVPR} & LSTM   & Res101   &  78.35 & \textbf{83.14} & 71.32 & \textbf{68.09} & \textbf{73.65}  & \textbf{58.03} & \textbf{67.99} &  \textbf{68.67}  & - &  5.1   \\
			CMN~\cite{cmn} & LSTM   & VGG16  &  - &  71.03 & 65.77 & - & 54.32  & 47.76 & - &  -  & - &  - \\
			DDPN~\cite{Yu2018RethinkingDA} & LSTM   & Res101   &   76.8   &  80.1 &  72.4 & 64.8 & 70.5 & 54.1 & - & -   & - & -\\
			DGA~\cite{Yang_2019_ICCV_DGA} & BiLSTM   & VGG16  &   -   &  78.42 &  65.53 & -  & 69.07 & 51.99 & - & 63.28 & - & 3\\    
			PLAN~\cite{Zhuang_2018_CVPR} & LSTM   & VGG16   &  - &  75.31 & 65.52 & - & 61.34  & 50.86 & - & -  & - & -\\
			RvGTree~\cite{RvGTree} & BiLSTM   & Res101   &  75.06 &  78.61 & 69.85 & 63.51 & 67.45  & 56.66 & 66.95 & 66.51  & - & -\\
			NMTREE~\cite{Liu_2019_ICCV} & BiLSTM   & Res101   &  76.41  &  81.21 & 70.09 & 66.46 & 72.02 & 57.52 & 65.87 & 66.44 & 106.5 & -\\
			\midrule
			\multicolumn{13}{l}{\textit{\textbf{Two-Stage} Vision-Language methods based on \textbf{Bert-base} for Language Model}} \\ 
			\midrule
			Uniter~\cite{uniter}  & Bert-B & Res101  & 81.24 & 86.48 & 73.94 & 75.31 & 81.3 & 65.58 & 74.31 & 74.51 & 190.3 & 8.1\\
			VLBERT~\cite{vlbert}  & Bert-B & Res101 & - & - & - & 71.6 & 77.72 & 60.99 & -	& - & 155.1 & 10.5\\
			VILLA~\cite{villa} & Bert-B & Res101 & \textbf{81.65} & \textbf{87.4} & \textbf{74.48} & \textbf{76.05} & \textbf{81.65} & \textbf{65.7} & \textbf{75.9} & 75.93 & 191.5 & 8.3\\
			ERNIE ViLL~\cite{ERNIEViL} & Bert-B  & Res101 & - & - & - & 74.02 & 80.33 & 64.74 & - & -  & - & -\\
			12 in 1~\cite{12in1} & Bert-B  & ResXT152  & - & \multicolumn{2}{c|}{80.58} & - & \multicolumn{2}{c|}{73.25} & - & \textbf{75.96} & - & - \\
			ViLBERT~\cite{vilbert} & Bert-B  & Res101  & -  & -  & -  & 72.34 &  78.52 & 62.61 & -  & - & 364 & 6 \\
			
			\midrule 
			\multicolumn{13}{l}{\textit{\textbf{Multi-Stage} Vision-Language methods} - Encoder-Decoder Transformers.} \\
			\multicolumn{13}{l}{*\textbf{Note}: these approaches are more powerful than encoder-only models.} 
			\\
			\midrule
			VGTR~\cite{vgtr} & Bi-LSTM & Res50  & 78.29 & 81.49 & 72.38 & 63.29 & 70.01 & 55.64 & 64.19 & 64.01 & - & -\\
			VGTR~\cite{vgtr} & Bi-LSTM & Res101  & 79.20 & 82.32 & 73.78 & 63.91 & 70.09 & 56.51 & 65.73 & 67.23 & - & -\\
			VL-T5~\cite{vlt5} & T5 & Res101 & - & - & - & - & - & - & - & 71.3 & 304.9 & 6.7 \\
			VLT~\cite{vltIccv2021} & RNN & Res50 & 76.20 & 80.31 & 71.44 & 64.19 & 68.40 & 55.84 & 61.03 & 60.24 & - & - \\
			MDETR~\cite{mdetr} & Roberta-B & Res101 & 86.75 & 89.58 & 81.41 & 79.52 & 84.09 & 70.62 & 81.64 & 80.89 & 185.2 & 12.24 \\
            MDETR~\cite{mdetr} & Roberta-B & ENB3 & \textbf{87.51} & \textbf{90.4} & \textbf{82.67} & \textbf{81.13} & \textbf{85.52} & \textbf{72.96} & \textbf{83.35} & \textbf{83.31} & 152.6 & 12.57 \\ 
			\midrule 
			\multicolumn{13}{l}{\textit{\textbf{Single-Stage} Vision-Language methods}} \\
			\midrule
			FAOA~\cite{Yang_2019_ICCV} & Bert & Darknet53  & 72.05  & 74.81 & 67.91 & -  & - & -  & - & - & 182.9 & 18.9\\
			RCCF~\cite{rccf} & BiLSTM   & DLA34  &   -   &  81.06 &  71.85 & - & 70.35 & 56.32 & - & - & - & 40\\
			SSG~\cite{ssg} & BiLSTM   & Darknet53   &   -   &  76.51 &  67.50 & - & 62.14 & 49.27 & 58.80 & - & - & -\\
			MCN~\cite{mcn} & BiGRU   & Darknet53  &   80.08 &  82.29 & 74.98  & 67.16 & 72.86 & 57.31  & 66.46 & 66.01 & - & -\\
			ReSC~\cite{resc} & Bert-B  & Darknet53  & 76.59 & 78.22 & 73.25 & 63.23 & 66.64 & 55.53 & 64.87 & 64.87 & 179.9 & 15.8\\
			TransVG~\cite{transvg} & Bert-B  & Res101  & 81.02 & 82.72 & \textbf{78.35} & 64.82 & 70.70 & 56.94 & 68.67 & 67.73 & 149.7 & 18.1\\
			\textbf{\modelname}  & Bert-B  & Linear  & \textbf{82.9} & \textbf{85.6} & 77.4 & \textbf{73.5} & \textbf{78.6} & \textbf{64.9} & \textbf{73.4} & \textbf{74.3} & \textbf{114.3} & \textbf{41.9}\\
			\midrule
		\end{tabular}
	}
	\label{tab:ref_reuslt_appendix}
\end{table}

\begin{table}
\centering
\caption{Ablation w/ and w/o pre-training.}
\scalebox{0.9}{
\begin{tabular}{|c|ccc|cc|}
\hline
 &  \multicolumn{3}{c|}{RefCoco} & \multicolumn{2}{c|}{CopsRef} \\
Pretraining &  val & testA & testB & val & test \\
\hline
w/o &  73.4 & 78 & 67.1 & 64.4 & 69.5\\
w/  &  82.9 \textcolor{forestgreen}{(+9.5)}  & 85.6 \textcolor{forestgreen}{(+7.6)} & 77.4 \textcolor{forestgreen}{(+10.3)} & 68.08 \textcolor{forestgreen}{(+3.68)} & 71.3 \textcolor{forestgreen}{(+1.8)} \\
\hline
\end{tabular}
}
\label{tab:ablade_pretraining}
\end{table}

\clearpage
\section{Ablation on Model Pretraining}
As mentioned in the main paper, \modelname is pre-trained on the concatenated detection dataset curated by \cite{mdetr} to allow a good initialization of the detection branch. Table~\ref{tab:ablade_pretraining} further highlights the importance of the pre-training on RefCoco and CopsRef dataset. The averaged gains are 9.13\% and 2.74\% on RefCoco and CopsRef respectively. This supports the need of pre-trainng stage for the detection branch.

\end{document}